\documentclass[11pt]{article}
\usepackage{acl2012}
\usepackage{times}
\usepackage{latexsym}
\usepackage{amsmath}
\usepackage[hyphens]{url}
\usepackage{breakurl} 
\makeatletter
\newcommand{\@BIBLABEL}{\@emptybiblabel}
\newcommand{\@emptybiblabel}[1]{}
\makeatother
\usepackage[breaklinks]{hyperref}

\usepackage{color}
\usepackage{pbox}
\usepackage{makecell}
\usepackage{verbatim} % allows multiline comments
\usepackage{subfigure} % for the subfloat in the figs
\usepackage{graphicx}
\usepackage{epsfig}
\usepackage{amssymb}
\usepackage{balance}
\usepackage{tablefootnote} % for footnotes in table
\usepackage{textcomp} % for \texttildelow

\usepackage{paralist} % for compact enum environment
\renewenvironment{itemize}[1]{\begin{compactitem}#1}{\end{compactitem}}
\renewenvironment{enumerate}[1]{\begin{compactenum}#1}{\end{compactenum}}

\usepackage{booktabs} % for tables, bottom rule etc
\newcommand{\hide}[1]{}
\newcommand{\xhdr}[1]{\vspace{1.7mm}\noindent{{\bf #1.}}}

\newcommand{\etc}{\emph{etc.}}
\newcommand{\eg}{\emph{e.g.}}
\newcommand{\ie}{\emph{i.e.}}

\usepackage{etoolbox}
\newcommand{\zerodisplayskips}{%
  \setlength{\abovedisplayskip}{4pt} %try 2pt
  \setlength{\belowdisplayskip}{4pt}
  \setlength{\abovedisplayshortskip}{4pt}
  \setlength{\belowdisplayshortskip}{4pt}}
\appto{\normalsize}{\zerodisplayskips}
\appto{\small}{\zerodisplayskips}
\appto{\footnotesize}{\zerodisplayskips}

\let\oldenumerate\enumerate
\renewcommand{\enumerate}{
  \vspace{-0.7\topsep} % removing space at top
  \oldenumerate
  \setlength{\itemsep}{1pt}
  \setlength{\parskip}{0pt}
  \setlength{\parsep}{0pt}
  \setlength{\topsep}{0pt}
  \setlength{\partopsep}{0pt}
}
\let\olditemize\itemize
\renewcommand{\itemize}{
  \vspace{-0.7\topsep}
  \olditemize
  \setlength{\itemsep}{1pt}
  \setlength{\parskip}{0pt}
  \setlength{\parsep}{0pt}
  \setlength{\topsep}{0pt}
  \setlength{\partopsep}{0pt}
}

\graphicspath{{./FIG/}}

\title{Large-scale Analysis of Counseling Conversations:\\An Application of Natural Language Processing to Mental Health}
\author{Tim Althoff\footnote[1]{},\;\; Kevin Clark\footnote[1]{},\;\; Jure Leskovec \\
Stanford University\\
  {\tt \{althoff, kevclark, jure\}@cs.stanford.edu}
  }

\begin{document}
\maketitle

\begin{abstract}
% !TEX root = paper-counseling.tex

% \todo{Use more successful and less successful counselors instead of good/bad!}
% \todo{Update figures to same visual language}
% \todo{Goal: TACL, Dec 1}
% \todo{make sure sections address these main results very clearly}

Mental illness is one of the most pressing public health issues of our time. 
While counseling and psychotherapy can be effective treatments,
our knowledge about how to conduct successful counseling conversations has been limited
due to lack of large-scale data with labeled outcomes of the conversations.
%because previous studies are largely qualitative and small-scale.
% How should we talk to others to help them feel better?
In this paper, we present a large-scale, quantitative study on the discourse of text-message-based counseling conversations. 
% (largest study published to date).
%% can we somehow include the general applicability.
%While some aspects of counseling are difficult to quantify, as a first step 
We develop a set of novel computational discourse analysis methods %suited for large-scale datasets 
to measure %analyze 
how various linguistic aspects of conversations are correlated with conversation outcomes. 
%While some aspects of counseling remain difficult to quantify, we apply... and discover...
Applying techniques such as 
sequence-based conversation models, 
language model comparisons,
%differences in language models, 
message clustering, and psycholinguistics-inspired word frequency analyses, %  to crisis counseling
we discover actionable conversation strategies that are associated with better conversation outcomes.
\end{abstract}

% \vspace{1mm}
% \noindent {\bf Categories and Subject Descriptors:} H.2.8 {\bf
% [Database Management]}: Database applications---{\it Data mining}

% \noindent {\bf General Terms:} Algorithms; Experimentation.

% \noindent {\bf Keywords:} Donor Retention; User Retention; Crowdfunding. 

%% anonymous submission
\renewcommand{\thefootnote}{\fnsymbol{footnote}}
\footnotetext[1]{Both authors contributed equally to the paper.}
\renewcommand{\thefootnote}{\arabic{footnote}}

% \jure{One general criticism of the paper is the flow. Somehow we nowhere really explain the outline and the questions we will be asking. Another problem is that most sections simply start with a couple of seemingly ``random'' questions and then the impression is that the paper jumps around and that there is no bigger structure to the paper. We have to do a better job giving a high-level flow of the paper and make sure that the transitions between the sections are smoother and better motivated.} 

\section{Introduction}
\label{sec:intro}
% !TEX root = paper-counseling.tex

% Lots of stats http://www.nami.org/Learn-More/Mental-Health-By-the-Numbers
Mental illness is a major global health issue. 
In the U.S. alone, 43.6 million adults (18.1\%) experience mental illness in a given year~\cite{mental_health_stats_1}. 
In addition to the person directly experiencing a mental illness, family, friends, and communities are also affected~\cite{insel2008assessing}.
% Serious mental illness costs America \$193.2 billion in lost earnings per year~\cite{insel2008assessing}.
% Suicide is the second leading cause of death for people aged 15--24~\cite{mental_health_stats_2}.
% More than 90\% of children who die by suicide have a mental health condition~\cite{us1999mental,mental_health_stats_2}.

In many cases, mental health conditions can be treated effectively through psychotherapy and counseling~\cite{mental_health_treatable}.
\enlargethispage{2\baselineskip} % otherwise too long
% (WHO, 2015)\nocite{mental_health_treatable}. %~\cite{mental_health_treatable}.
However, it is far from obvious how to best conduct counseling conversations.
Such conversations are free-form without strict rules, and involve many choices that could make a difference in someone's life.
%How should one talk to others to help them feel better?
Thus far, quantitative evidence for effective conversation strategies
%data on what really helps 
has been scarce, since most studies on counseling have been limited to very small sample sizes and qualitative observations (\eg, Labov and Fanshel, (1977); Haberstroh et al., (2007)\nocite{labov1977therapeutic,haberstroh2007experience}).
% This has lead many counselors to conduct their conversations in whatever way they find to work best. \todo{find ref}
%
However, recent advances in technology-mediated counseling conducted online or through texting~\cite{haberstroh2007experience} have allowed counseling services to scale with increasing demands and to collect large-scale data on counseling conversations and their outcomes. % ,crisistextline2015mainpage
% Such data is ideally suited for linguistic discourse analysis since everything is observed about the interaction between the counselor and the conversation partner. 
% Recently, advances in web technology have allowed technology-mediated counseling conducted online or through texting~\cite{crisistextline2015mainpage}.
% These advances have allowed to scale counseling services to meet the increasing demand and to collect data on what counseling conversations have yielded positive outcomes and which ones did not.
% %
% Such data is ideally suited for linguistic discourse analysis since everything is observed about the interaction between the counselor and the conversation partner. 
% For example, all conversations share a clear and common goal, the outcome of whether or not this goal was achieved is observed (through follow-up surveys), and we are not messing non-textual dynamics such as mimicry or pitch.

Here we present the largest study on counseling conversation strategies published to date. % (to the best of our knowledge). 
% \jure{One sentence about the setting: Done or wanted?}
We use data from an SMS texting-based counseling service where people in crisis (depression, self-harm, suicidal thoughts, anxiety,  %relationship, 
\etc), engage in therapeutic conversations with counselors. 
%We 
%develop novel computational discourse analysis methods to 
%analyze 
The data contains millions of messages from eighty thousand counseling conversations conducted by hundreds of counselors over the course of one year. %We 
%\footnote{We refer to crisis counselors as \textit{counselors} for the remainder of the paper for brevity.}
%While in general counseling is very complex, we find that
 % we find that  counselors exhibit a large variety of behaviors, we find 
%there are significant quantifiable differences between more successful and less successful counselors in how they conduct conversations (Section~\ref{sec:counseling_quality}).
%
We develop a set of computational methods suited for large-scale discourse analysis to study how various linguistic aspects of conversations are correlated with conversation outcomes (collected via a follow-up survey). 
%While in general counseling is very complex, 

% copied from section 4.
% There are several benefits of focusing analyses on counselors (rather than individual conversations): First, we are interested in general conversation strategies rather than properties of main issues (\eg, depression vs. suicide). 
% While each conversation is different and will revolve around its main issue, we assume that counselors have a particular style and strategy that is invariant across conversations.
% % (and has developed during through training and experience).
% % Conducting the analysis on conversation level across many issues can also address this claim but conflates different strategies by different actors.
% %
% Second, we assume that conversation quality is noisy. Even a very good counselor will face some hard conversations in which they do everything right but are still unable to make their conversation partner feel better.
% Over time, however, the ``true'' quality of the counselor will become apparent.
% %
% Third, our goal is to understand successful conversation strategies and to make use of these insights in counselor training.
% Focusing on the counselor is helpful in understanding, monitoring, and improving counselors' conversation strategies.

We focus our analyses on counselors instead of individual conversations because 
we are interested in general conversation strategies rather than properties of specific issues.
%that can be used in counselor training 
We find that 
% we find that counselors exhibit a large variety of behaviors, we find 
there are significant, quantifiable differences between more successful and less successful counselors in how they conduct conversations. 
%

% Adding this as glue/outline.
% \jure{CUT THIS SINCE IT IS REPETITIVE: First, we show that successful counselors use more varied language than less successful counselors (Section~\ref{sec:counselor_adaptability})
% and then describe several aspects of this variation 
% in terms of language content (Section~\ref{sec:ambiguity})
% and in terms of how counselors lead the conversation through different stages (Section~\ref{sec:conversation_progress}). 
% Then, we study the effect on the conversation partner in terms of perspective change (Section~\ref{sec:perspective_change}).}

Our findings suggest actionable strategies that are associated with successful counseling:
% \pagebreak[4]
\begin{enumerate}
  \item \textbf{Adaptability (Section~\ref{sec:counselor_adaptability}):} Measuring the distance between vector representations of the language used in conversations going well and going badly, we find that successful counselors are more sensitive to the current trajectory of the conversation and react accordingly.
  \item \textbf{Dealing with Ambiguity (Section~\ref{sec:ambiguity}):} We develop a clustering-based method to measure differences in how counselors respond to very similar ambiguous situations. We learn that successful counselors clarify situations by writing more, reflect back to check understanding, and make their conversation partner feel more comfortable through affirmation.
  \item \textbf{Creativity (Section~\ref{subsec:templates_and_creativity}):} We quantify the diversity in counselor language by measuring cluster density in the space of counselor responses 
  %  in the space of counselor responses
  %Examining the cluster density in the space of all counselor questions, 
  and find that successful counselors respond in a more creative way, not copying the person in distress exactly and not using too generic or ``templated'' responses. 
  \item \textbf{Making Progress (Section~\ref{sec:conversation_progress}):} We develop a novel sequence-based unsupervised conversation model able to discover ordered conversation stages common to all conversations. Analyzing the progression of stages, we determine that successful counselors are quicker to get to know the core issue and faster to move on to collaboratively solving the problem.
  \item \textbf{Change in Perspective (Section~\ref{sec:perspective_change}):} We develop novel measures of perspective change using psycholinguistics-inspired word frequency analysis. We find that people in distress are more likely to be more positive, think about the future, and consider others, when the counselors bring up these concepts. We further show that this perspective change is associated with better conversation outcomes consistent with psychological theories of depression.
  %Measuring coordination between the conversation partners we find that texters are more likely to be more positive, think about the future, and consider others as well if the counselors brings up these concepts. 
  %Through psycholingustics-inspired word frequency analysis, we find that this perspective change is associated with better conversation outcomes consistent with psychological theories of depression (Section~\ref{sec:perspective_change}).
\end{enumerate}  
Further, we demonstrate that counseling success on the level of individual conversations is predictable using features based on our discovered conversation strategies (Section~\ref{sec:prediction}). 
% \jure{Such predictive tools could be used to help counselors better progress through the conversation and would result in better counseling.}
Such predictive tools could be used to help counselors better progress through the conversation and could result in better counseling practices.
The dataset used in this work has been released publicly and more information on dataset access can be found at \url{http://snap.stanford.edu/counseling}.

Although we focus on crisis counseling in this work, our proposed methods more generally apply to other conversational settings and can be used to study how language in conversations relates to conversation outcomes.

\section{Related Work}
\label{sec:related_work}
% !TEX root = paper-counseling.tex

% \todo{
%   hierarchical bootstrapping: We use the member bootstrapping technique from \cite{ren2010nonparametric}. 
%   Add this to first analysis/plot that uses it!
% }

% In this section, we review related work from therapeutic discourse analysis and psycholinguistics
% as well as large-scale computational linguistics.

Our work relates to two lines of research:

% \jure{We could do a better job adding a sentence to each section and saying how we relate to the work. Right now the related work is a laundry list without making any relation to our research work.}

\xhdr{Therapeutic Discourse Analysis \& Psycholinguistics}
The field of conversation analysis was born in the 1960s out of a suicide prevention center~\cite{sacks1995lectures,van1997discourse}. Since then conversation analysis has been applied to various clinical settings including psychotherapy~\cite{labov1977therapeutic}.
% The field of conversation analysis was born out of a suicide prevention center in Los Angeles in the 1960s~\cite{sacks1995lectures,van1997discourse}.
% Since then, therapeutic discourse has been a focus of a
% wide range of clinical settings, including psychotherapy~\cite{labov1977therapeutic}.
%
Work in psycholinguistics has demonstrated that the words people use %in their daily lives 
can reveal important aspects of their social and psychological worlds~\cite{pennebaker2003psychological}.
% Previous work has found 
%that confronting upsetting experiences through writing is associated with positive health outcomes~\cite{pennebaker1988disclosure}, and 
Previous work also found that there are linguistic cues associated with depression~\cite{ramirez2008psychology,campbell2003secret} as well as with suicude~\cite{pestian2012sentiment}. 
%such as using negatively connotated words and first person pronouns
% there are linguistic cues associated with depression that can be measured computationally~\cite{ramirez2008psychology},
% essays written by depressed participants used more negatively valenced words and used the word ``I'' more than did non-depressed participants,
% and that participants who always write in first person are less likely to get better over time~\cite{campbell2003secret}.
These findings are consistent with Beck's\nocite{beck1967depression} cognitive model of depression (1967; cognitive symptoms of depression precede the affective and mood symptoms)
% which posits that
and with Pyszczynski and Greenberg's\nocite{pyszczynski1987self} self-focus model of depression (1987; depressed persons engage in higher levels of self-focus than non-depressed persons).
% which posits that

%
% Lastly, a qualitative study of the experiences of 5 participants concluded that online counseling provides a promising and practical, therapeutic alternative or adjunctive resource to face-to-face counseling for some populations~\cite{haberstroh2007experience}.
% Online counseling was found to be a promising practical alternative to face-to-face counseling~\cite{haberstroh2007experience}.
%
In this work, we propose an operationalized psycholinguistic model of perspective change 
and further provide empirical evidence for these theoretical models of depression.

\xhdr{Large-scale Computational Linguistics Applied to Conversations}
Large-scale studies have revealed subtle dynamics in conversations such as 
coordination or style matching effects~\cite{niederhoffer2002linguistic,Danescu-Niculescu-Mizil:thesis2012} as well as expressions of social power and status~\cite{bramsen2011extracting,Danescu-Niculescu-Mizil+al:12a}.
Other studies have connected writing to measures of success in the context of requests~\cite{althoff2014howtoaskforafavor}, 
user retention~\cite{althoff2015donor}, 
novels~\cite{ashok2013success}, and scientific abstracts~\cite{guerini2012do}.
% \jure{THIS PART DOES NOT CONNECT: Maybe just take these two citations and add them to the previous sentence.}
% and connected writing style to success for novels~\cite{ashok2013success} and scientific abstracts~\cite{guerini2012do}.
%
% and that the level of accommodation is related to power and status~\cite{Danescu-Niculescu-Mizil+al:12a}. 
% of the conversation partners
% Further, changes in language over time have also been used as indicators for behaviors such as leaving an online community~\cite{danescu2013no}.
%
% \xhdr{Conversation Modeling}
Prior work has modeled dialogue acts in conversational speech based on linguistic cues and discourse coherence~\cite{stolcke2000dialogue}.
Unsupervised machine learning models have also been used to model conversations and segment them into speech acts, topical clusters, or stages. 
Most approaches employ Hidden Markov Model-like models~\cite{barzilay2004catching,ritter2010unsupervised,paul2012mixed,YanMcALesPenSha14} 
% In this work, we use a model similar to~\cite{YanMcALesPenSha14} to find progression stages within each conversation.
which are also used in this work to model progression through conversation stages.
%

% now counseling stuff
Very recently, technology-mediated counseling has allowed the collection of large datasets on counseling.
Howes et al. (2014)\nocite{howes2014linguistic} find that symptom severity can be predicted from transcript data with comparable accuracy to face-to-face data but suggest that insights into style and dialogue structure are needed to predict measures of patient progress.
Counseling datasets have also been used to predict the conversation outcome~\cite{ElvaMasterThesis:2015} but without modeling the within-conversation dynamics that are studied in this work. %using features inspired by psycholinguistics
% In contrast to the present work, researchers have not explored dynamics within conversations.
Other work has explored how novel interfaces based on topic models can support counselors  during conversations~(Dinakar et al., 2014a; 2014b; 2015; Chen, 2014).\nocite{dinakar2014real,dinakar2014stacked,dinakar2015mixed,GeChenMasterThesis:2014}
% ~\cite{dinakar2014real,dinakar2014stacked,dinakar2015mixed,GeChenMasterThesis:2014}.

Our work joins these two lines of research by developing computational discourse analysis methods applicable to large datasets that are grounded in therapeutic discourse analysis and psycholinguistics.

\section{Dataset Description} % and Quality?
\label{sec:dataset}
% !TEX root = paper-counseling.tex

% \subsection{Counseling conversations}

% % The textual nature enables conversation analysis at large scale and with clear annotation of conversation outcomes (\eg, compared to audio recordings that would have to be transcribed and cleaned).
% We obtained access to such conversations through a partnership with an organization that provides support for people via text message (henceforth referred to as \textit{organization})\footnote{Our dataset will be publicly released through {ANONYMIZED URL}.}. %\url{http://www.crisistextline.org/open-data/}
% % Approval to perform conversation analysis was obtained from the IRB at Stanford University (protocol \#33809).
% Approval to perform conversation analysis was obtained from the IRB at Anonymized University (protocol \#12345).

%\xhdr{Texting-based Crisis Counseling}
In this work, we study anonymized counseling conversations from 
%Crisis Text Line (CTL), 
a not-for-profit organization providing free crisis intervention via SMS messages. Text-based counseling conversations are particularly well suited for conversation analysis because all interactions between the two dialogue partners are fully observed (\ie, there are no non-textual or non-verbal cues). Moreover, the conversations are important, constrained to dialogue between two people, and outcomes can be clearly defined (\ie, we follow up with the conversation partner as to whether they feel better afterwards), which enables the study of how conversation features are associated with actual outcomes.

\xhdr{Counseling Process}
Any person in distress can text the organization's public number. Incoming requests are put into a queue and an available counselor picks the request from the queue and engages with the incoming conversation. We refer to the crisis counselor as the \textit{counselor} and the person in distress as the {\em texter}. After the conversation ends, the texter receives a follow-up question (``How are you feeling now? Better, same, or worse?'') % as well as a reminder that they can text CTL again.
which we use as our conversation quality ground-truth (we use binary labels: good versus same/worse, since we care about improving the situation).
% which is meaningful since our goal is to make texter actually feel better).
% other reasons: balance of responses, look similar, simplicity
In contrast to previous work that has used human judges to rate a caller's crisis state~\cite{kalafat2007evaluation}, we directly obtain this feedback from the texter.
Furthermore, the counselor fills out a post-conversation report (\eg, suicide risk, main issue such as depression, relationship, self-harm, suicide, \etc). All crisis counselors receive extensive training and commit to weekly shifts for a full year.
%About 20\% of all texters reply to the follow up question. %$31.7\% acc. to Noelle;  for us 0.192
% Asking the texter directly is arguably superior to more indirect alternatives (\eg, asking human judges to rate and compare caller’s crisis state at the beginning and at the end of their calls as in~\cite{kalafat2007evaluation}).

\begin{table}[t]
\centering
\small
\begin{tabular}{ l l }
Dataset statistics & \\ \hline
Conversations & 80,885 \\
Conversations with survey response & 15,555 (19.2\%) \\
Messages & 3.2 million \\ 
Messages with survey response & 663,026 (20.6\%)\\ 
Counselors & 408 \\ 
Messages per conversation* & 42.6 \\
Words per message* & 19.2
\end{tabular}
\vspace{-3mm}
\caption{Basic dataset statistics. Rows marked with * are computed over conversations with survey responses.}
% \todo{Or maybe just report over all conversations? Numbers would look pretty much the same}}
\label{tab:stats}
\end{table}

\xhdr{Dataset Statistics} Our dataset contains 408 counselors and %3,215,568 
3.2 million messages in 80,885 conversations between November 2013 and November 2014 (see Table~\ref{tab:stats}).
% (about one message every ten seconds).
All system messages (\eg, instructions), as well as texts that contain survey responses (revealing the ground-truth label for the conversation) were filtered out.
Out of these conversations, we use the 15,555, or 19.2\%, that contain a ground-truth label (whether the texter feels better or the same/worse after the conversation) for the following analyses. Conversations span a variety of issues of different difficulties (see rows one and two of Table~\ref{tab:issues}).
Approval to analyze the dataset was obtained from the Stanford IRB.
%
% The dataset has been released publicly and more information on dataset access can be found at \url{http://snap.stanford.edu/counseling}.
% \\ {\small \url{http://snap.stanford.edu/counseling}}.
%\todo{While the clear supervised setting is important for this work, future work should investigate how to deal with the weakly or unsupervised setting and how to increase response rates.}

\begin{table*}[t]
\centering
\resizebox{2.1\columnwidth}{!}{%
\footnotesize 
\setlength{\tabcolsep}{5pt}
\begin{tabular}{ l c c c c c c c c c } 
%Issue $\longrightarrow$
    & NA &  Depressed  & Relationship & Self harm & Family & Suicide & Stress & Anxiety & Other \\ \hline
   Success rate & 0.556 & 0.612 & 0.659 & 0.672 & 0.711 & 0.573 & 0.696 & 0.671 & 0.537   \\ %\hline
   Frequency &  0.200 & 0.200 & 0.089 & 0.074 & 0.071 & 0.063 & 0.041 & 0.039 & 0.035   \\ %\hline
   \makecell[l]{Frequency with more \\successful counselors} & 0.203 & 0.199 & 0.089 & 0.067 & 0.072 & 0.061 & 0.048 & 0.042 & 0.030   \\ %\hline
   \makecell[l]{Frequency with less \\successful counselors} &  0.223 & 0.208 & 0.087 & 0.070 & 0.067 & 0.056 & 0.030 & 0.032 & 0.028   
\end{tabular}
} %
\vspace{-3mm}
\caption{Frequencies and success rates for the nine most common conversation issues (NA: Not available). On average, more and less successful counselors face the same distribution of issues. 
%The distributions of issues facing more and less successful counselors are similar.
}
\label{tab:issues}
\end{table*}

% Our dataset contains 4,298,687 messages in 93,064 conversations on the CTL platform between August 2013 and November 2014 (about one message every seven seconds).
% During the first months, CTL was not operating nationwide yet and the data show several platform tests, so we only use conversations between November 2013 and November 2014 (\ie, one year).
% We further filter all system messages, as well as texts that contain survey responses (and would reveal the ground-truth label for the conversation).
% This leaves us with 3,215,568 messages in 80,885 conversations.
% Out of those conversations, 15,555 or 19.2\% contain a ground-truth label (texter feels same/better/worse after conversation).
% We use these for the following analysis.

\section{Defining Counseling Quality}
\label{sec:counseling_quality}
% !TEX root = paper-counseling.tex

The primary goal of this paper is to study strategies that lead to conversations with positive outcomes. Thus, we require a ground-truth notion of conversation quality. 
In principle, we could study individual conversations and 
aim to understand what factors make the conversation partner (texter) feel better.
% afterwards
% aim to understand whether the conversation partner (texter) will feel better after a given conversation. 
However, it is advantageous to focus on the conversation actor (counselor) instead of individual conversations.

There are several benefits of focusing analyses on counselors (rather than individual conversations): First, we are interested in general conversation strategies rather than properties of main issues (\eg, depression vs. suicide). 
While each conversation is different and will revolve around its main issue, we assume that counselors have a particular style and strategy that is invariant across conversations.
% (and has developed during through training and experience).
% Conducting the analysis on conversation level across many issues can also address this claim but conflates different strategies by different actors.
%
Second, we assume that conversation quality is noisy. Even a very good counselor will face some hard conversations in which they do everything right but are still unable to make their conversation partner feel better.
Over time, however, the ``true'' quality of the counselor will become apparent.
Third, our goal is to understand successful conversation strategies and to make use of these insights in counselor training.
Focusing on the counselor is helpful in understanding, monitoring, and improving counselors' conversation strategies.

\begin{figure}[t]
  \centering
  \includegraphics[width=.95\linewidth]{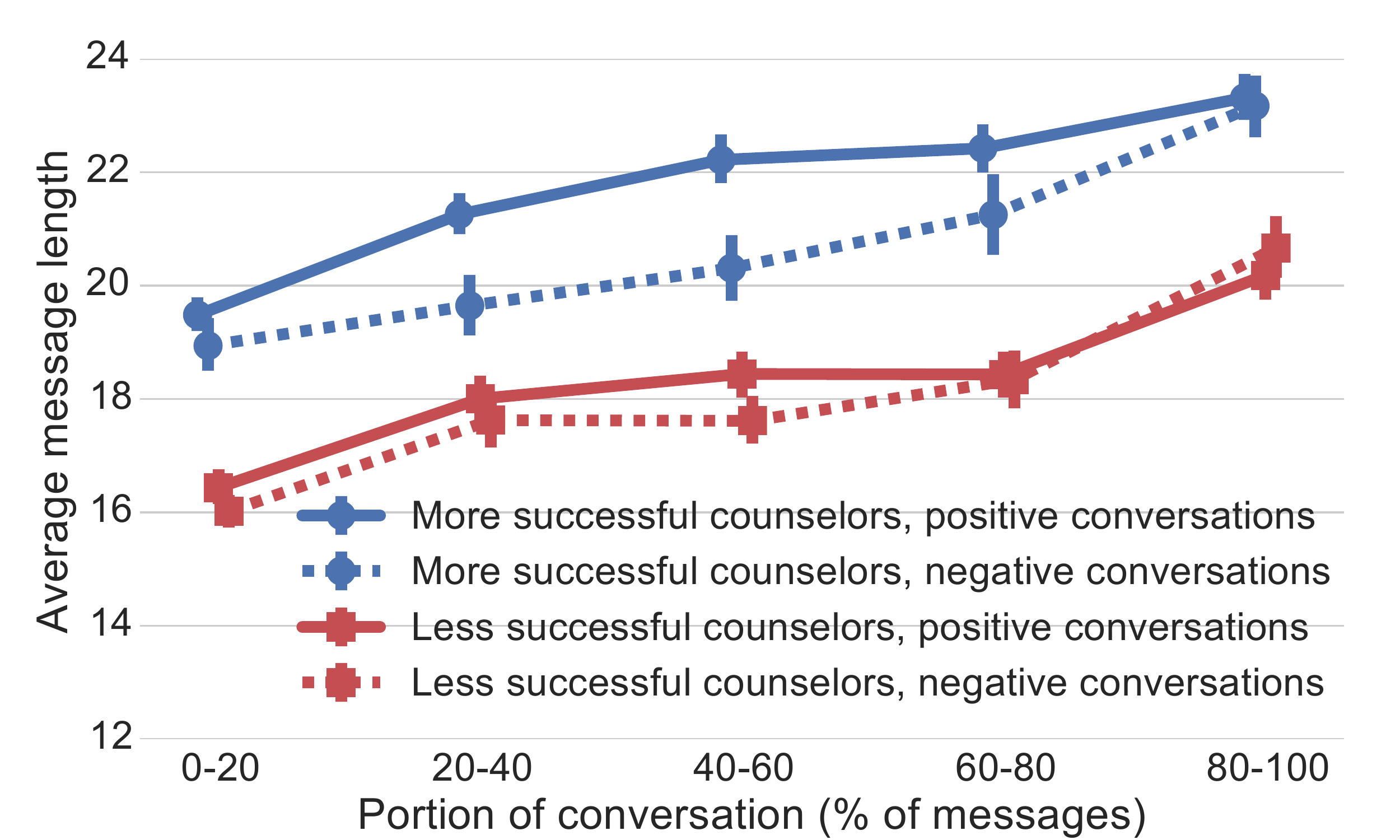}
  % \vspace{-2mm}
  \caption{%Differences in counselor message length over the course of the conversation are not between positive/negative conversations (solid/dashed) but more and less successful counselors (blue circle/red square).\\
  Differences in counselor message length (in \#tokens) over the course of the conversation are larger between more and less successful counselors (blue circle/red square) than between positive and negative conversations (solid/dashed).
  Error bars in all plots correspond to bootstrapped 95\% confidence intervals
  using the member bootstrapping technique from~\protect\newcite{ren2010nonparametric}. %accounting for the hierarchical nature of the data.
  % to account for the hierarchical nature of counselors and conversations. %of the corresponding mean estimate
  }
  \label{fig:message_length_counselor_crossproduct}
\end{figure}

\xhdr{More vs. Less Successful Counselors}
We split the counselors into two groups and then compare their behavior. Out of the 113 counselors with more than 15 labeled conversations of at least 30 messages each, we use the most successful 40 counselors as ``more successful'' counselors and the bottom 40 as ``less successful'' counselors. 
Their average success rates are 66.3-85.5\% and 42.1-58.6\%, respectively. % for the two groups,
%
% \xhdr{Positive vs. Negative Conversations}
While the counselor-level analysis is of primary concern, we will also differentiate between counselor behavior in ``positive'' versus ``negative'' conversations (\ie, those that will eventually make the texter feel better vs. not). Thus, in the remainder of the paper we differentiate between more vs. less successful counselors and positive vs. negative conversations. Studying the cross product of counselors and conversations allows us to gain insights on how both groups behave in positive and negative conversations. For example,  
Figure~\ref{fig:message_length_counselor_crossproduct} illustrates why differentiating between counselors and as well as conversations is necessary: differences in counselor message length over the course of the conversation are bigger between more and less successful counselors than between positive and negative conversations.
%
%An illustrative example for why differentiating between counselors and not only conversations is necessary is given in Figure~\ref{fig:message_length_counselor_crossproduct}. 
%The figure shows that differences in counselor message length over the course of the conversation are largely between more and less successful counselors, not positive and negative conversations.
%It shows that differences in counselor message length over the course of the conversation are not between positive/negative conversations but more and less successful counselors (which we would have missed otherwise).
% If we did not differentiate between good and bad counselors as well, we would have found that positive conversations have longer messages (which is true but the larger effect is between counselors).

\xhdr{Initial Analysis}
Before focusing on detailed analyses of counseling strategies we %quickly 
address two important questions: Do counselors specialize in certain issues? And, do successful counselors appear successful only because they handle ``easier'' cases?

To gain insights into the ``specialization hypothesis'' we make use the counselor annotation of the main issue (depression, self-harm, \etc).
We compare success rates of counselors across different issues and find that successful counselors have a higher fraction of positive conversations across all issues and that less successful counselors typically do not excel at a particular issue.
% \jure{Explain what we did and include some numbers (1 sentence). For example: We compare success rates of conversations across different issues and find that successful counselors have a higher fraction of positive conversations across all issues.}
Thus, we conclude that counseling quality is a general trait or skill and supporting that the split into more and less successful counselors is meaningful.
%\footnote{\label{fn:supplementary_material}Figure available in online supplementary material.}

Another simple explanation of the differences between more and less successful counselors could be that successful counselors simply pick ``easy'' issues.
%It could be that more successful counselors are only more successful because they are assigned ``easier'' conversations but we find that this is not the case.
However, we find that this is not the case. % (Table~\ref{tab:issues}).
In particular, we find that both counselor groups are very similar in how they select conversations from the queue (picking the top-most in 60.1\% vs. 60.3\%, respectively), work similar shifts, and handle a similar number of conversations simultaneously (1.98 vs. 1.83).
% Further, we do not find any significant differences between the two groups in terms of when they they work and how many conversations they handle simultaneously (1.98 vs 1.83).
% We do find that more successful counselors handle more conversations per shift but no evidence that shift length is correlated with conversation quality.
%
Further, we find that both groups face similar distributions of issues over time (see Table~\ref{tab:issues}). We attribute the largest difference, ``NA'' (main issue not reported), to the more successful counselors being more diligent in filling out the post-conversation report and having fewer conversations that end before the main issue is introduced.

\section{Counselor Adaptability}
\label{sec:counselor_adaptability}
% !TEX root = paper-counseling.tex

% \todo{All these results would have to show differences between good and less successful counselors in good/bad conversation (that's what they adapt to)}

In the remainder of the paper we focus on factors that mediate the outcome of a conversation. First, we examine whether successful counselors are more aware that their current conversation is going well or badly and study how the counselor adapts to the situation.
We investigate this question by looking for language differences between positive and negative conversations. In particular, we compute a distance measure between the language counselors use in positive conversations and the language counselors use in negative conversations and observe how this distance changes over time. 

We capture the time dimension by breaking up each conversation into five even chunks of messages. Then, for each set of counselors (more successful or less successful), conversation outcome (positive or negative), and chunk (first 20\%, second 20\%, etc.), we build a TF-IDF vector of word occurrences to represent the language of counselors within this subset. 
We use the global inverse document (\ie, conversation) frequencies instead of the ones from each subset to make the vectors directly comparable and control for different counselors having different numbers of conversations by weighting conversations so all counselors have equal contributions. We then measure the difference between the ``positive'' and ``negative'' vector representations by taking the cosine distance in the induced vector space. We also explored using Jensen-Shannon divergence between traditional probabilistic language models %instead of TF-IDF representations 
and found these methods gave similar results.%, although they required careful smoothing to account for rare words.
%If we measured the counselor's behavior in some very narrow way (\eg, how often the counselor uses a particular word or phrase), we would never be able to show the absence of behavioral differences. 
%Therefore we propose a very general methodology capturing \textit{language variability} instead.
%Since all behavior (except for timing differences) is exclusively textual we can quantify opengeneral behavioral differences by quantifying the distance between language models of counselors in good versus bad conversations.

\begin{figure}[t]
  \centering
  \includegraphics[width=.95\linewidth]{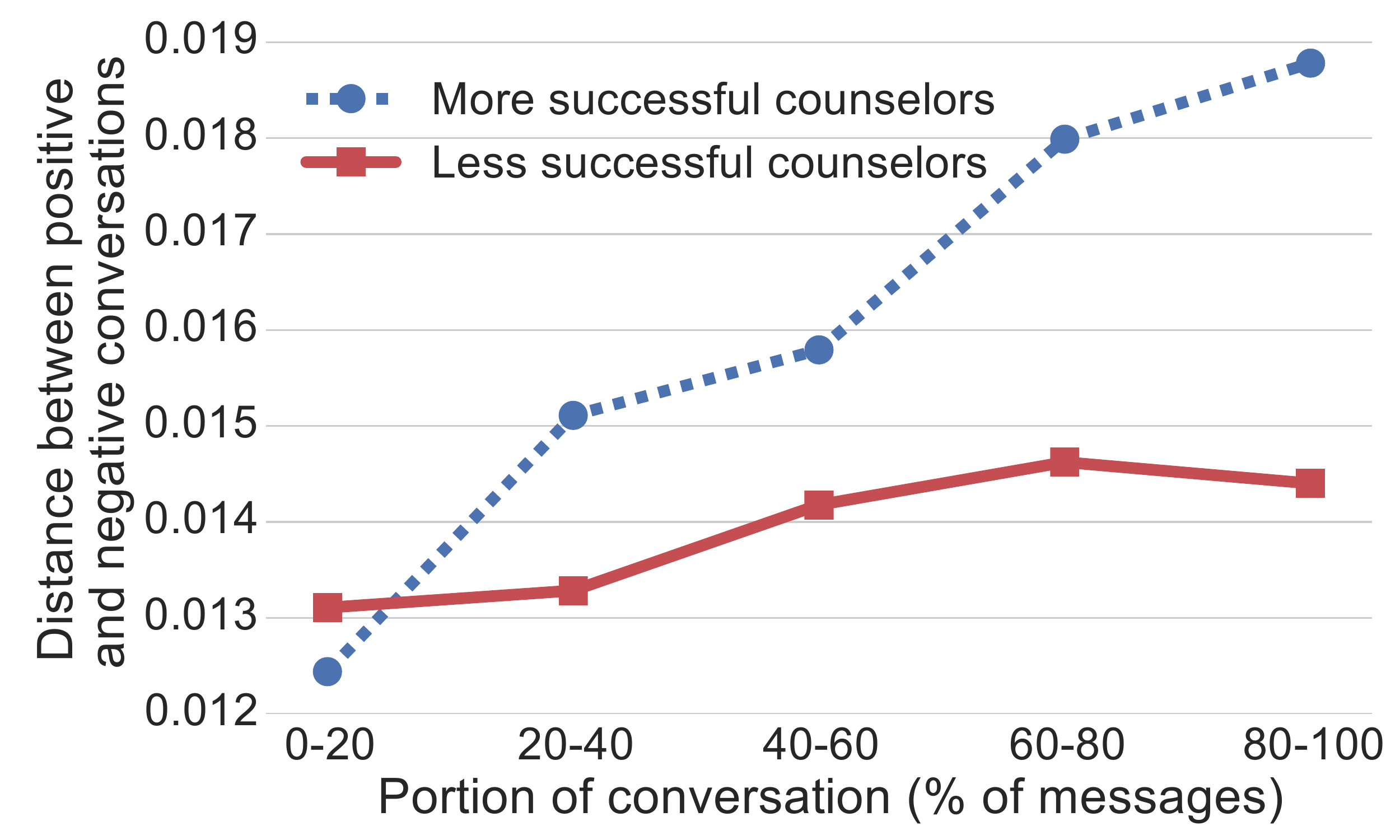}
  \vspace{-4mm}
  \caption{More successful counselors are more varied in their language across positive/negative conversations, suggesting they adapt more. All differences between more successful and less successful counselors except for the 0-20 bucket were found to be statistically significant ($p<0.05$; bootstrap resampling test).}
  \label{fig:positive_negative_language_similarity}
\end{figure}

\xhdr{Results}
We find more successful counselors are more sensitive to whether the conversation is going well or badly and vary their language accordingly 
% adapt more to how the conversation is going 
(Figure~\ref{fig:positive_negative_language_similarity}). 
At the beginning of the conversation, the language between positive and negative conversations is quite similar, %, as the counselors have little information about how well the conversation is going. 
but then the distance in language increases over time.
% However, the distance in language increases over time.
% as the conversations start going well or poorly and counselors react. 
This increase in distance is much larger for more successful counselors than less successful ones,
%The distance for more successful counselors over time is larger compared to less successful counselors, 
% We found this difference increases much more for more successful counselors than unsuccessful ones, 
suggesting they are more aware of when conversations are going poorly and 
adapt their counseling more in an attempt to remedy the situation. %All differences except for the 0-20 bucket were found to be statistically significant $(p<0.05)$ according to a bootstrap resampling test. 

\section{Reacting to Ambiguity}
\label{sec:ambiguity}
% !TEX root = paper-counseling.tex

Observing that successful counselors are better at adapting to the conversation, we next examine 
\textit{how} counselors differ and what factors determine the differences.
% what features differentiate between successful and less successful counselors.
In particular, domain experts have suggested that more successful counselors are better at handling ambiguity in the conversation~\cite{levitt2005promoting}.
Here, we use \textit{ambiguity} to refer to the uncertainty of the situation and the texter's actual core issue resulting from insufficiently short or uncertain descriptions.
% http://onlinelibrary.wiley.com/doi/10.1002/j.2164-490X.2005.tb00055.x/abstract
% When a someone brings up an issue, particularly one that is hard to talk about, it often takes a while for the heart of the issue to be revealed.
Does initial ambiguity of the situation negatively affect the conversation?
How do more successful counselors deal with ambiguous situations?  

\xhdr{Ambiguity}
Throughout this section we measure ambiguity in the conversation as the shortness of the texter's responses in number of words.
% We study the level of initial ambiguity in the conversation and measure ambiguity as the shortness of the texter's responses.
While ambiguity could also be measured through concreteness ratings of the words in each message (\eg, using concreteness ratings from~\newcite{brysbaert2014concreteness}), 
we find that results are very similar and that length and concreteness are strongly related and hard to distinguish.

\begin{figure}[t]
  \centering
  \includegraphics[width=.90\linewidth]{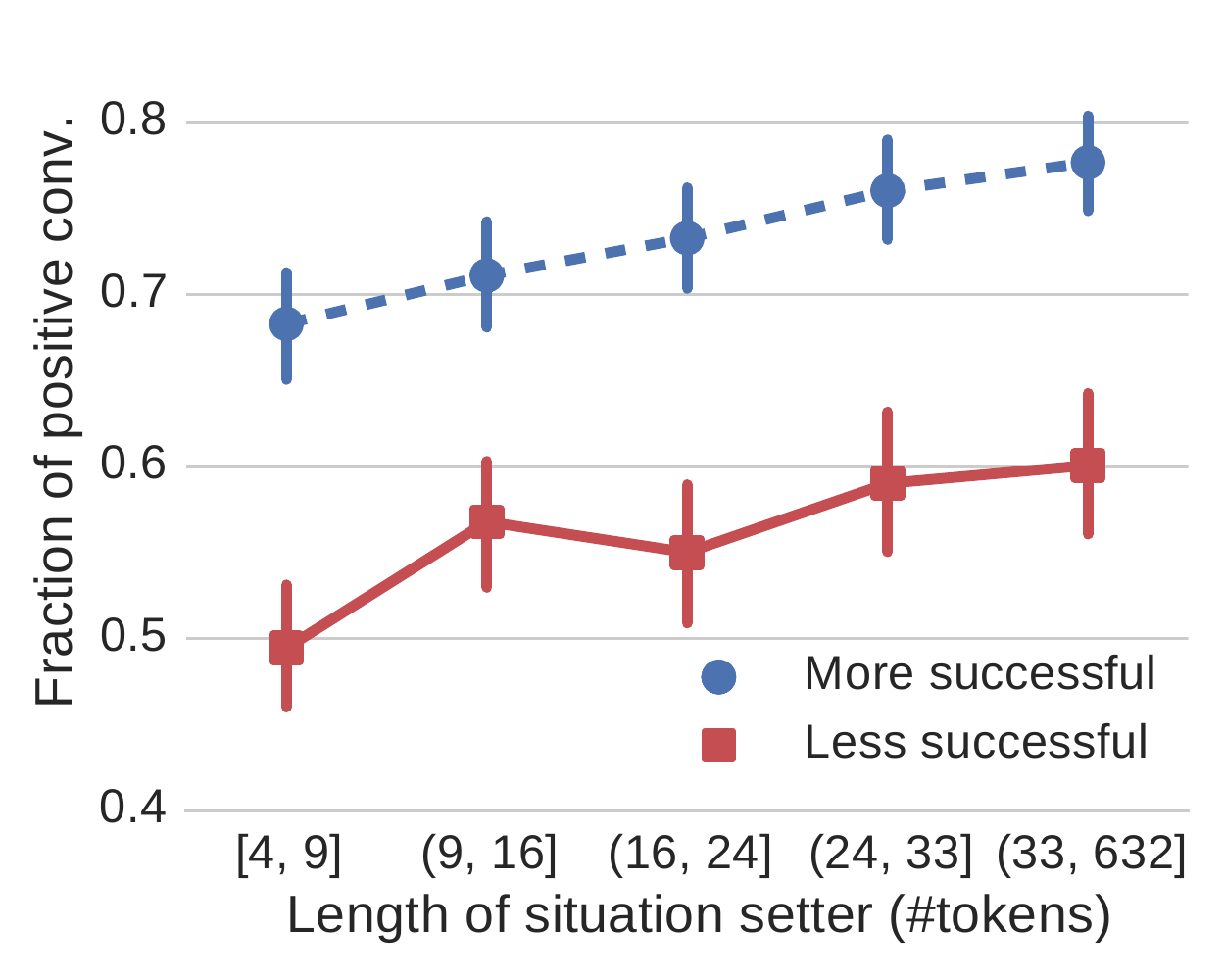}
  \vspace{-3mm}
  \caption{More ambiguous situations (length of situation setter) are less likely to result in positive conversations.}
  \label{fig:conv_quality_by_situation_length}
\end{figure}

\begin{figure}[t]
  \centering
  \includegraphics[width=.90\linewidth]{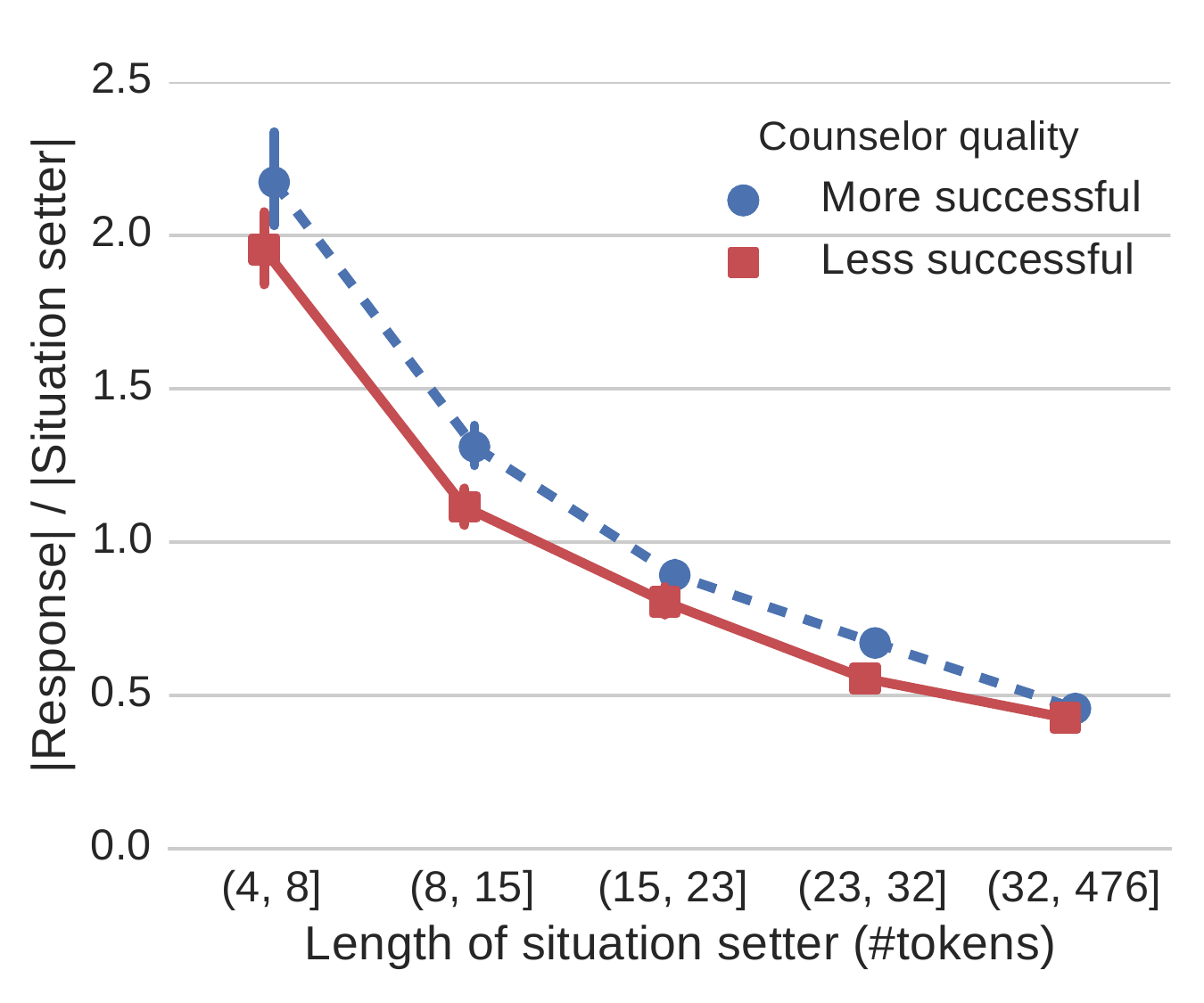}
  \vspace{-3mm}
  \caption{All counselors react to short, ambiguous messages by writing more (relative to the texter message) but more successful counselors do it more than less successful counselors.
  }
  \label{fig:length_counselor_reaction_to_short_situation_setter}
\end{figure}

\subsection{Initial Ambiguity and Situation Setter}
\label{subsec:situation_setter_ambiguity}

%
%In our corpus counselors almost always react with a question, but what makes a good question in an ambiguous situation?
%\xhdr{Method}
% Answering these questions in general is challenging since it strongly depends on conversation context (\eg, all earlier messages and questions).
It is challenging to measure ambiguity and reactions to ambiguity at arbitrary points throughout the conversation since it strongly depends on the context of the entire conversation (\ie, all earlier messages and questions).
However, we can study nearly identical \textit{beginnings} of conversations where we can directly compare how more successful and less successful counselors react given nearly identical situations (the texter first sharing their reason for texting in).
%(and where ambiguity is largest).
%
% We study how responses of more successful and less successful counselors vary within nearly identical situations (the texter first sharing their reason for texting in) and the effect on the next texter message as well as the conversation as a whole.
We identify the \textit{situation setter} within each conversation as the first long message by the texter (typically a response to a ``Can you tell me more about what is going on?'' question by the counselor).

\xhdr{Results}
We find that ambiguity plays an important role in counseling conversations. 
Figure~\ref{fig:conv_quality_by_situation_length} shows that more ambiguous situations (shorter length of situation setter) are less likely to result in successful conversations (we obtain similar results when measuring concreteness~\cite{brysbaert2014concreteness} directly).
% instead of message length).
% It is possible the driving factor is the unwillingness by the texter to share more rather than the 'symptom' of short messages.
Further, we find that counselors generally react to short and ambiguous situation setters by writing significantly more than the texters (Figure~\ref{fig:length_counselor_reaction_to_short_situation_setter}; if counselors wrote exactly as much as the texter, we would expect a horizontal line $y=1$).
However, more successful counselors react more strongly to ambiguous situations than less successful counselors.

% \todo{if we want to bring up word vectors - there is some discussion in the comments}
% discussion of embedding models for this (we tried embedding models to capture the state of the conversation/situation with mixed success. 
% word vectors work well for synonyms, and e.g. first names are all treated quite differently.
% however, some words need more context (killing time, killing myself, this is killing me), 
% domain: and some words have particular meaning in our context (cutting means almost exclusively self-harm, but not in most other corpora)
% similar: boyfriend and dad appear in similar lexical contexts but it might often be a different issue?
% also, a few words in each message carry most of the meaning (e.g cutting, suicide). 
% Here IDF has worked better to weight some terms more strongly. (and can easily be applied to bigrams)

% \subsection{What Do Good Questions Look Like?}
%\subsection{What is a Good Response to Ambiguity?}
\subsection{How to Respond to Ambiguity}
\label{subsec:respond_to_ambiguity}

%Are there differences in how more and less successful counselors respond to nearly identical situations? What effect does this have on the conversation?
Having observed that ambiguity plays an important role in counseling conversations, we now examine in greater detail how counselors respond to nearly identical situations.

%\xhdr{Approac}
We match situation setters by representing them through TF-IDF vectors on bigrams and find similar situation setters as nearest neighbors within a certain cosine distance in the induced space.\footnote{
  Threshold manually set after qualitative analysis of matches from randomly chosen clusters. 
  Results were not overly sensitive to threshold choice, choice of representation (\eg, word vectors), and distance measure (\eg, Euclidean).
}
We only consider situation setters that are part of a dense cluster with at least 10 neighbors, allowing us to compare follow-up responses by the counselors
(4829/12770 situation setters were part of one of 589 such clusters).
% we only consider long conv with at  least X messages in conv
We also used distributed word embeddings (\eg, \cite{mikolov2013distributed}) instead of TF-IDF vectors but found the latter to produce better clusters. % (likely due to better term weighting).  % experimented with 

Based on counselor training materials we hypothesize that more successful counselors
\begin{itemize}
\item address ambiguity by writing more themselves, 
\item use more check questions (statements that tell the conversation partner that you understand them while avoiding the introduction of any opinion or advice~\cite{labov1977therapeutic}; \eg ``that sounds like...''),
% -> wertschaetzung
% e.g. https://echopen.wordpress.com/2011/05/04/reflective-listening/
\item check for suicidal thoughts early (\eg, ``want to die''), % early
\item thank the texter for showing the courage to talk to them (\eg, ``appreciate''),
\item use more hedges (mitigating words used to lessen the impact of an utterance; \eg, ``maybe'', ``fairly''),
\item and that they are less likely to respond with surprise (\eg, ``oh, this sounds really awful'').
\end{itemize}
A set of regular expressions is used to detect each class of responses (similar to the examples above).

% Based on counselor training materials we hypothesize that %~\cite{crisistextline2015trainingmaterials}
% more successful counselors might deal with ambiguity by writing more themselves,
% using more check questions (statements that tell the conversation partner that you understand them while avoiding the introduction of any opinion or advice~\cite{labov1977therapeutic}; \eg ``that sounds like...''),
% % -> wertschaetzung
% % e.g. https://echopen.wordpress.com/2011/05/04/reflective-listening/
% checking for suicidal thoughts early (\eg, ``want to die''),
% thanking the texter for showing the courage to talk to them (\eg, ``appreciate''),
% using more hedges (mitigating words used to lessen the impact of an utterance; \eg, ``maybe'', ``fairly''),
% and that they would be less likely to respond with surprise (\eg, ``oh, this sounds really awful'').
% A set of regular expressions is used to detect each class of responses (similar to the examples above).
% \todo{empathy, surprise...}
% see https://en.wikipedia.org/wiki/Hedge_(linguistics)
% we use what Cristian was using in crisis paper
% TODO do counselors write more if texter writes less?

\xhdr{Results}
We find several statistically significant differences in how counselors respond to nearly identical situation setters 
(see Table~\ref{tab:situation_setter_matching_results}). %; weighing all clusters equally).
% On average, every cluster of situation setters contains around five responses by more successful counselors as well as less successful counselors.
%We find that situations handled by more successful counselors are significantly longer and more often result in a positive conversation.
While situation setters tend to be slightly longer for more successful counselors (suggesting that conversations are not perfectly randomly assigned), counselor responses are significantly longer and also spur longer texter responses.
Further, the more successful counselors respond in a way that is less similar to the original situation setter (measured by cosine similarity in TF-IDF space) compared to less successful counselors (but the texter's response does not seem affected).
% This could mean that they are able to steer the conversation more and could also be affected by what they use their initial reaction for.
%
%
We do find that more successful counselors use more check questions, check for suicide ideation more often, show the texter more appreciation, and use more hedges,
but we did not find a significant difference with respect to responding with surprise.

\begin{table}[t]
\centering
\resizebox{1.0\columnwidth}{!}{%
\begin{tabular}{lccc}
 &  More S. &  Less S. & Test\\
 \hline
\% conversations successful & 70.7 & 51.7 & ***\\
\#messages in conversation & 57.0 & 46.7 & ***\\
Situation setter length (\#tokens) & 12.1 & 10.7 & ***\\ % \tablefootnote{For a discussion of how conversations are assigned to counselors refer to Section~\ref{sec:counseling_quality}.}
C response length (\#tokens) & 15.8 & 11.8 & ***\\
T response length (\#tokens) & 20.4 & 18.8 & ***\\
\% Cosine sim. C resp. to context & 11.9 & 14.8 & ***\\
\% Cosine sim. T resp. to context & 7.6 & 7.3 & --\\
\% C resp. w check question & 12.6 & 4.1 & ***\\
\% C resp. w suicide check & 13.5 & 10.3 & ***\\
\% C resp. w thanks & 6.3 & 2.4 & ***\\
\% C resp. w hedges & 41.4 & 36.8 & ***\\
\% C resp. w surprise & 3.3 & 2.8 & --
\end{tabular}
}
% \vspace{-5mm}
\caption{
Differences between more and less successful counselors (C; More S. and Less S.) in responses to nearly identical situation setters (Sec.~\ref{subsec:situation_setter_ambiguity}) by the texter (T).
Last column contains significance levels of Wilcoxon Signed Rank Tests (*** $p<0.001$, -- $p>0.05$).
}
\label{tab:situation_setter_matching_results}
\end{table}

\begin{figure}[t]
  \centering
  \includegraphics[width=.90\linewidth]{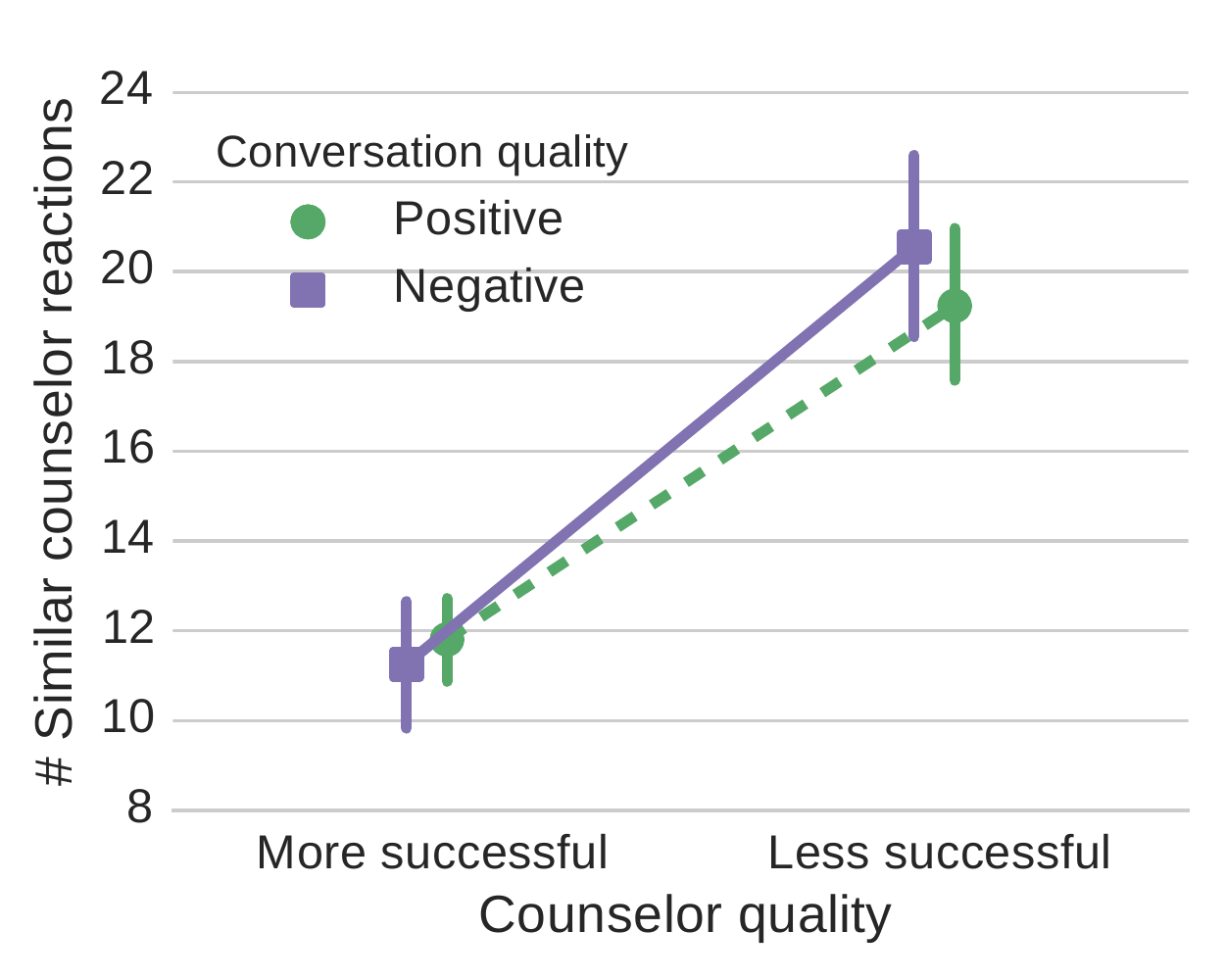}
  \vspace{-3mm}
  \caption{More successful counselors use less common/templated responses (after the texter first explains the situation). This suggests that they respond in a more creative way.
  There is no significant difference between positive and negative conversations.
  }
  \label{fig:counselor_common_reaction_by_counselor_quality}
\end{figure}

\subsection{Response Templates and Creativity}
\label{subsec:templates_and_creativity}
% \jure{Why 6.2. suggests that counselors should stick to predefined templates?}
In Section~\ref{subsec:respond_to_ambiguity}, we observed that more successful counselors make use of certain templates (including check questions, checks for suicidal thoughts, affirmation, and using hedges).
While this could suggest that counselors should stick to such predefined templates, 
we find that, in fact, more successful counselors do respond in more creative ways.

% While the results in Section~\ref{subsec:respond_to_ambiguity} could suggest that counselors should stick to predefined templates (including check questions, checks for suicidal thoughts, affirmation, and using hedges),
% we find that more successful counselors respond in a more creative way.

%\xhdr{Method}
We define a measure of how ``templated'' the counselors responses are by counting the number of similar responses in TF-IDF space for the counselor reaction (\textit{c.f.}, Section~\ref{subsec:respond_to_ambiguity}; again using a manually defined and validated threshold on cosine distance).

%\xhdr{Results}
Figure~\ref{fig:counselor_common_reaction_by_counselor_quality} shows that more successful counselors use less common/templated questions.
% again this result seems robust wrt what questions we look at, what space we use, and what exact thresholds
% This suggests that they adapt more towards the case at hand.
This suggests that while more successful counselors questions follow certain patterns, they are more \textit{creative} in their response to each situation.
This tailoring of responses requires more effort from the counselor, which is consistent with the results in Figure~\ref{fig:message_length_counselor_crossproduct} that showed that more successful counselors put in more effort in composing longer messages as well.

% \todo{could include brief comment on question words: have some results on that W questions better than yes/no questions (19 vs 16 token responses on avg)}

% \todo{Talk about check questions instead of ``tentafiers''.
% (from dans book)
% ``The speech act we will look at, introduced very briefly above, is often called a
% check or a check question (Labov and Fanshel 1977, Carletta et al. 1997b). A
% check is a subtype of question which requests the interlocutor to confirm
% some information; the information may have been mentioned explicitly in the
% preceding dialogue (as in the example below), or it may have been inferred
% from what the interlocutor has said.''
% }

\section{Ensuring Conversation Progress}
\label{sec:conversation_progress}
% !TEX root = paper-counseling.tex

%\begin{itemize}
%  \item ``saying the right things at the right time'' (right now only time chunked language model, or time varying feature plots?)
%  \item HMM-like stage model
%  \item perspective change (in progress) (1. self vs others, 2. past/present/future) \todo{for this to show up here more successful counselors would have to be better at facilitating it}
%  \item Description of HMM model: TODO
%  \item Even though we allow the model to jump forward and skip stages we find in practice that this almost never happens (makes sense since model makes use of full ``power'' this way...)
%  \item Result: more successful counselors are quicker to get to know texter and issue (stage 2) and use more of their time in the ``collaborative'' problem solving phase (stage 4). See Figure \ref{fig:stages_durations_good_bad}
%\end{itemize}

% Tim: I'm adding this as "glue".
After demonstrating content-level differences between counselors, we now explore temporal differences in how counselors progress through conversations.
% This section moves from content to temporal analysis and describes how conversations and counselors progress through several stages.
Using an unsupervised conversation model, we are able to discover distinct conversation stages
and find differences between counselors in how they move through these stages.
We further provide evidence that these differences could be related to power and authority by measuring linguistic coordination between the counselor and texter.
% First, we propose an unsupervised conversation model to discover conversation stages.
% Using this model, we find differences between counselors in how they make progress.
% Lastly, we explain these differences in terms of linguistic coordination and differences in power and authority.

% Most crisis counseling conversations have a similar high-level structure as counselors first introduce themselves, get to know the texter and their situation, and then engage in problem solving by trying to come up with constructive ways in which the texter can cope with their situation. In fact, the counselors are trained to follow specific guidelines on how to move a conversation forward. To capture this structure, we employ unsupervised conversation modeling techniques to explore how counselors move between different stages of the conversation. 

\subsection{Unsupervised Conversation Model}
Counseling conversations follow a common structure due to the nature of conversation as well as counselor training.
Typically, counselors first introduce themselves, get to know the texter and their situation, and then engage in constructive problem solving.
We employ unsupervised conversation modeling techniques to capture this stage-like structure within conversations.

\begin{figure}[t]
  \centering
  \includegraphics[width=.75\linewidth]{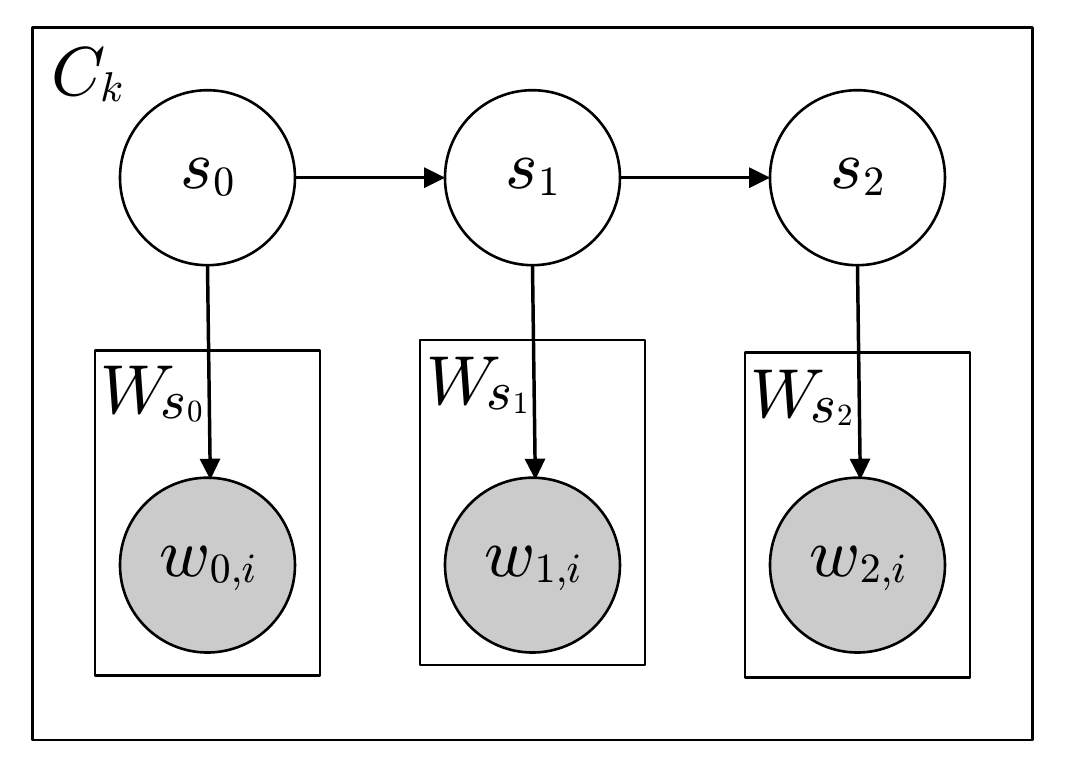}
  \vspace{-4mm}
  \caption{Our conversation model generates a particular conversation $C_k$ by first generating a sequence of hidden states $s_0, s_1, ...$ according to a Markov model. Each state $s_i$ then generates a message as a bag of words $w_{i, 0}, w_{i, 1}, ...$ according a unigram language model $W_{s_i}$.}
  \label{fig:hmm_model}
\end{figure}

\begin{figure}[t]
  \centering
  \includegraphics[width=.90\linewidth]{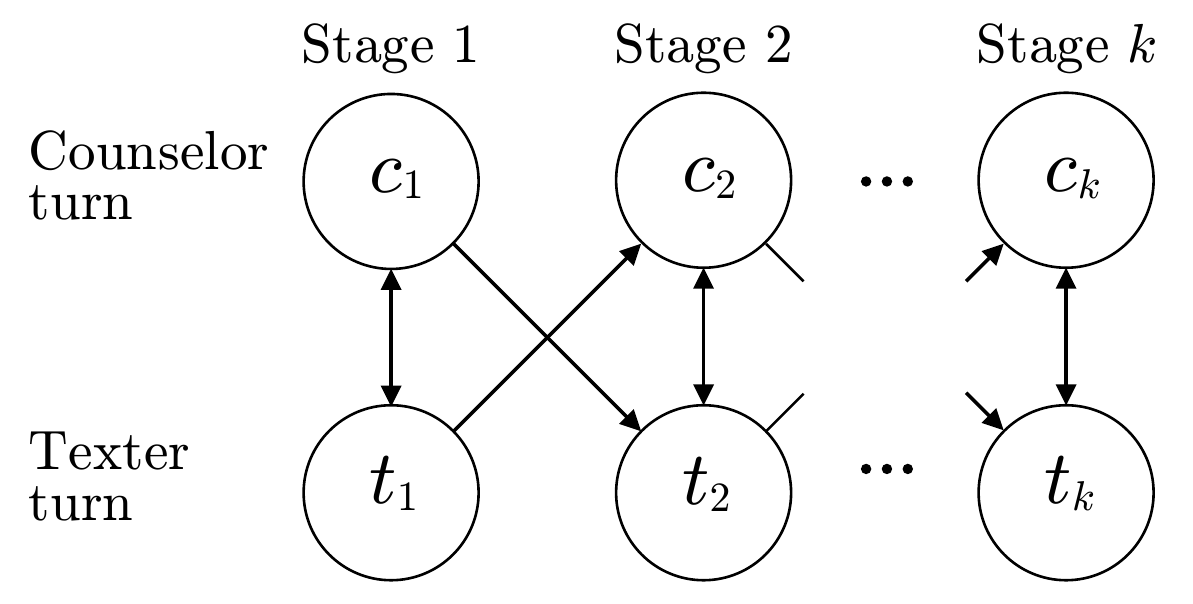}
  \vspace{-3mm}
  \caption{Allowed state transitions for the conversation model. Counselor and texter messages are produced by distinct states and conversations must progress through the stages in increasing order.}
  \label{fig:hmm_transitions}
\end{figure}

%\todo{Justin: explain what this model is. also notation in fig 8 is not explained.}
Our conversation model is a message-level Hidden Markov Model (HMM). %inspired by Ritter et al. (2010)\nocite{ritter2010unsupervised}.
Figure~\ref{fig:hmm_model} illustrates the basic model where hidden states of the HMM represent \textit{conversation stages}. % instead of topics.
%We extend prior work by imposing a fixed ordering on the stages and only allowing transitions from the current stage to the same stage or a later one. 
Unlike in prior work on conversation modeling, we impose a fixed ordering on the stages and only allow transitions from the current stage to the next one (Figure~\ref{fig:hmm_transitions}).
This causes it to learn a fixed dialogue structure common to all of the counseling sessions  %(similar to~\cite{YanMcALesPenSha14}) 
as opposed to conversation topics. 
% Similar approaches have been employed to identify progression stages in sequences of events \cite{YanMcALesPenSha14}.
Furthermore, we separately model counselor and texter messages by treating their turns in the conversation as distinct states. We train the conversation model with expectation maximization, using the forward-backward algorithm to produce the distributions during each expectation step. We initialized the model with each stage producing messages according to a unigram distribution estimated from all messages in the dataset and uniform transition probabilities. The unigram language models are defined over all words occurring more than 20 times (over 98\% of words in the dataset), with other words replaced by an unknown token.

\begin{table*}[ht]
\centering
\small % \footnotesize
 \resizebox{2.0\columnwidth}{!}{%
\begin{tabular}{llll}
  % \hline
   Stage & Interpretation & Top words for texter & Top words for counselor \\ \hline
  1 & Introductions & hi, hello, name, listen, hey & hi, name, hello, hey, brings \\ 
  2 & Problem introduction & dating, moved, date, liked, ended & gosh, terrible, hurtful, painful, ago \\ 
  3 & Problem exploration & knows, worry, burden, teacher, group & react, cares, considered, supportive, wants \\ 
  4 & Problem solving & write, writing, music, reading, play & hobbies, writing, activities, distract, music \\ 
  5 & Wrap up & goodnight, bye, thank, thanks, appreciate & goodnight, 247, anytime, luck, 24 \\ %\hline
\end{tabular}
} %
\vspace{-2mm}
\caption{The top 5 words for counselors and texters with greatest increase in likelihood of appearing in each stage. 
The model successfully identifies interpretable stages consistent with counseling guidelines (qualitative interpretation based on stage assignment and model parameters; only words occurring more than five hundred times are shown). %in the stage
}
\label{tab:stages_words}
\end{table*}

\begin{figure}[t]
  \centering
  \vspace{-2mm}
  \includegraphics[width=1.0\linewidth]{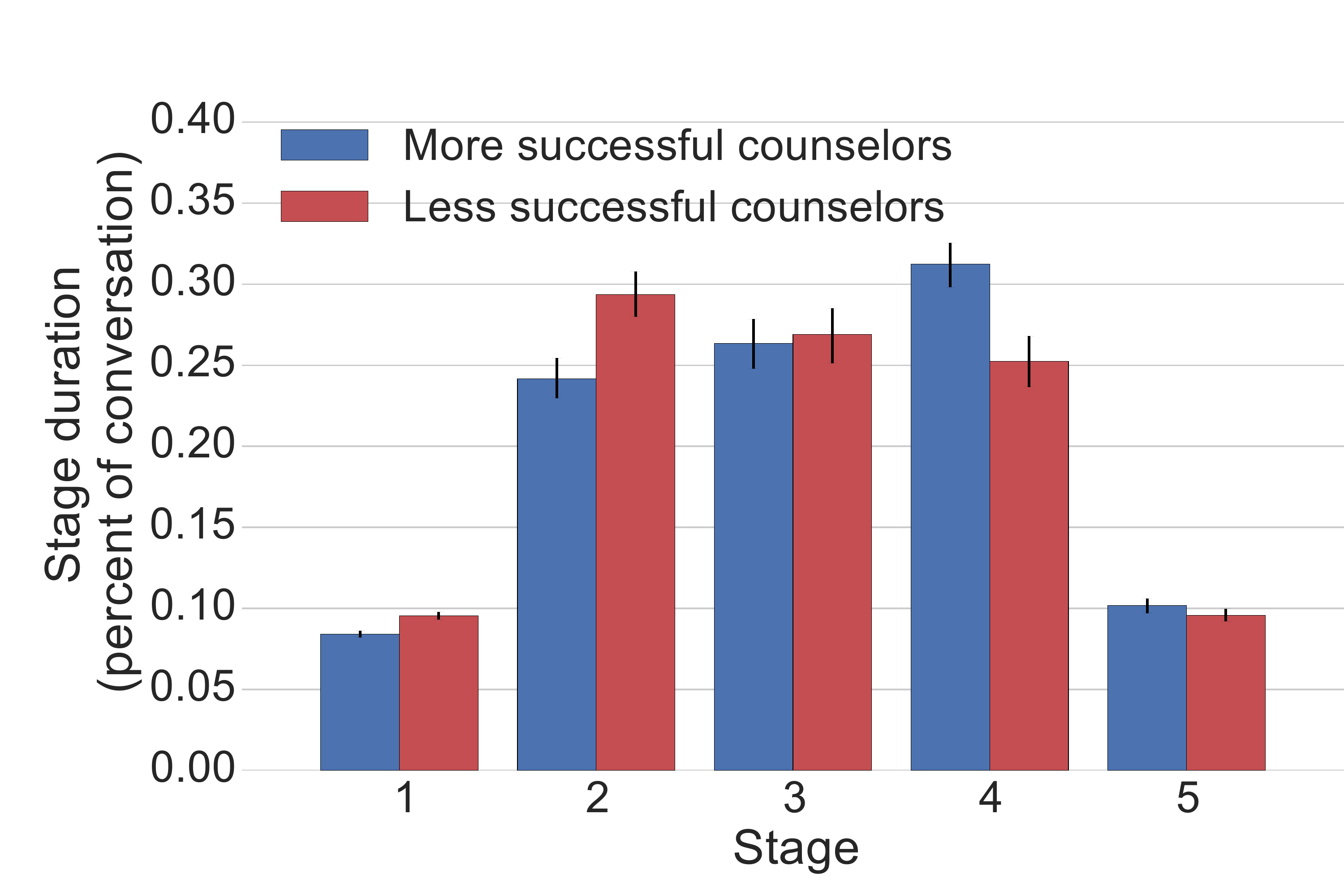}
  \vspace{-10mm}
  \caption{More successful counselors are quicker to get to know texter and issue (stage 2) and use more of their time in the ``problem solving'' phase (stage 4).}
  \label{fig:stages_durations_good_bad}
\end{figure}

% After experimenting with different numbers of stages, 
\xhdr{Results}
We explored training the model with various numbers of stages and found five stages to produce a distinct and easily interpretable representation of a conversation's progress. Table~\ref{tab:stages_words} shows the words most unique to each stage. The first and last stages consist of the basic introductions and wrap-ups common to all conversations. In stage 2, the texter introduces the main issue, while the counselor asks for clarifications and expresses empathy for the situation. In stage 3, the counselor and texter discuss the problem, particularly in relation to the other people involved. 
In stage 4, the counselor and texter discuss actionable strategies that could help the texter.
This is a well-known part of crisis counselor training called ``collaborative problem solving."
% Crisis counselors call this stage ``collaborative problem solving," and train their counselors to bring up actionable solutions the texter can pursue after the conversation's conclusion. 

\subsection{Analyzing Counselor Progression}
Do counselors differ in how much time they spend at each stage?
In order to explore how counselors progress through the stages, we use the Viterbi algorithm to assign each conversation the most likely sequence of stages according to our conversation model. We then compute the average duration in messages of each stage for both more and less successful counselors.
We control for the different distributions of positive and negative conversations among more successful and less successful counselors by giving the two classes of conversations equal weight and 
control for different conversation lengths by only including conversations between 40 and 60 messages long. 

\xhdr{Results}
We find that more successful counselors are quicker to move past the earlier stages, particularly stage 2, and spend more time in later stages, particularly stage 4 (Figure~\ref{fig:stages_durations_good_bad}). 
This suggests they are able to more quickly get to know the texter and then spend more time in the problem solving phase of the conversation, which could be one of the reasons they are more successful.

\subsection{Coordination and Power Differences}
\label{subsec:coordination}
% \subsection{Explaining Differences through Linguistic Coordination}
% \xhdr{Coordination and Power Differences}
%More successful counselor's ability to move quickly to the productive ``collaborative problem solving'' stage suggests they are better at guiding the conversation forward. 
One possible explanation for the more successful counselors' ability to quickly move through the early stages is that they have more ``power'' in the conversation and can thus exert more control over the progression of the conversation.
We explore this idea by analyzing linguistic coordination, %in the counseling conversations
which measures how much the conversation partners adapt to each other's conversational styles. %unconsciously
Research has shown that conversation participants who have a greater position of power coordinate less (\emph{i.e.}, they do not adapt their linguistic style to mimic the other conversational participant as strongly) \cite{Danescu-Niculescu-Mizil+al:12a}. 

In our analysis, we use the ``Aggregated 2'' coordination measure $C(B, A)$ from \newcite{Danescu-Niculescu-Mizil:thesis2012}, which measures how much group $B$ coordinates to group $A$ (a higher number means more coordination). The measure is computed by counting how often specific markers (e.g., auxiliary verbs) are exhibited in conversations. 
If someone tends to use a particular marker right after their conversation partner uses that marker, it suggests they are coordinating to their partner. 

Formally, let set $S$ be a set of exchanges, each involving an initial utterance $u_1$ by $a \in A$ and a reply $u_2$ by $b \in B$. Then the coordination of $b$ to $A$ according to a linguistic marker $m$ is:
\[
C^m(b, A) = P(\mathcal{E}^m_{u_2 \to u_1} | \mathcal{E}^m_{u_1}) - P(\mathcal{E}^m_{u_2 \to u_1})
\]
where $\mathcal{E}^m_{u_1}$ is the event that utterance $u_1$ exhibits $m$ (\emph{i.e.}, contains a word from category $m$) and $\mathcal{E}^m_{u_2 \to u_1}$ is the event that reply $u_2$ to $u_1$ exhibits $m$. The probabilities are estimated across all exchanges in $S$. To aggregate across different markers, we average the coordination values of $C^m(b, A)$ over all markers $m$ to get a macro-average $C(b, A)$. The coordination between groups $B$ and $A$ is then defined as the mean of the coordinations of all members of group $B$ towards the group $A$.
%:
%\[
%C(B, A) = \langle C(b, A)\rangle_{b \in B} 
%\]

We use eight markers from Danescu-Niculescu-Mizil (2012)\nocite{Danescu-Niculescu-Mizil:thesis2012}, which are considered to be processed by humans in a generally non-conscious fashion: articles, auxiliary verbs, conjunctions, high-frequency adverbs, indefinite pronouns, personal pronouns, prepositions, and quantifiers.

%\todo{Justin: so is a higher number mean more coordination?}
%todo: go into the details of how these numbers are computed?
\xhdr{Results}
Texters coordinate less than counselors, with texters having a coordination value of $C(\text{texter, counselor})$=0.019 compared to the counselor's $C(\text{counselor, texter})$=0.030, suggesting that the texters hold more ``power'' in the conversation. 
However, more successful counselors coordinate less than less successful ones 
($C(\text{more succ. counselors, texter})$=0.029 vs. $C(\text{less succ. counselors, texter})$=0.032). 
All differences are statistically significant ($p < 0.01$; Mann-Whitney U test). 
This suggests that more successful counselors act with more control over the conversation, which could explain why they are quicker to make it through the initial conversation stages. 
% authority and control
% is perhaps one reason for their ability to quickly make progress through the initial conversation stages. 
%todo: the number is definitely lower the 0.05, double check

\section{Facilitating Perspective Change}
\label{sec:perspective_change}
% !TEX root = paper-counseling.tex

\begin{figure*}[t]
  \centering
  \includegraphics[width=.99\linewidth]{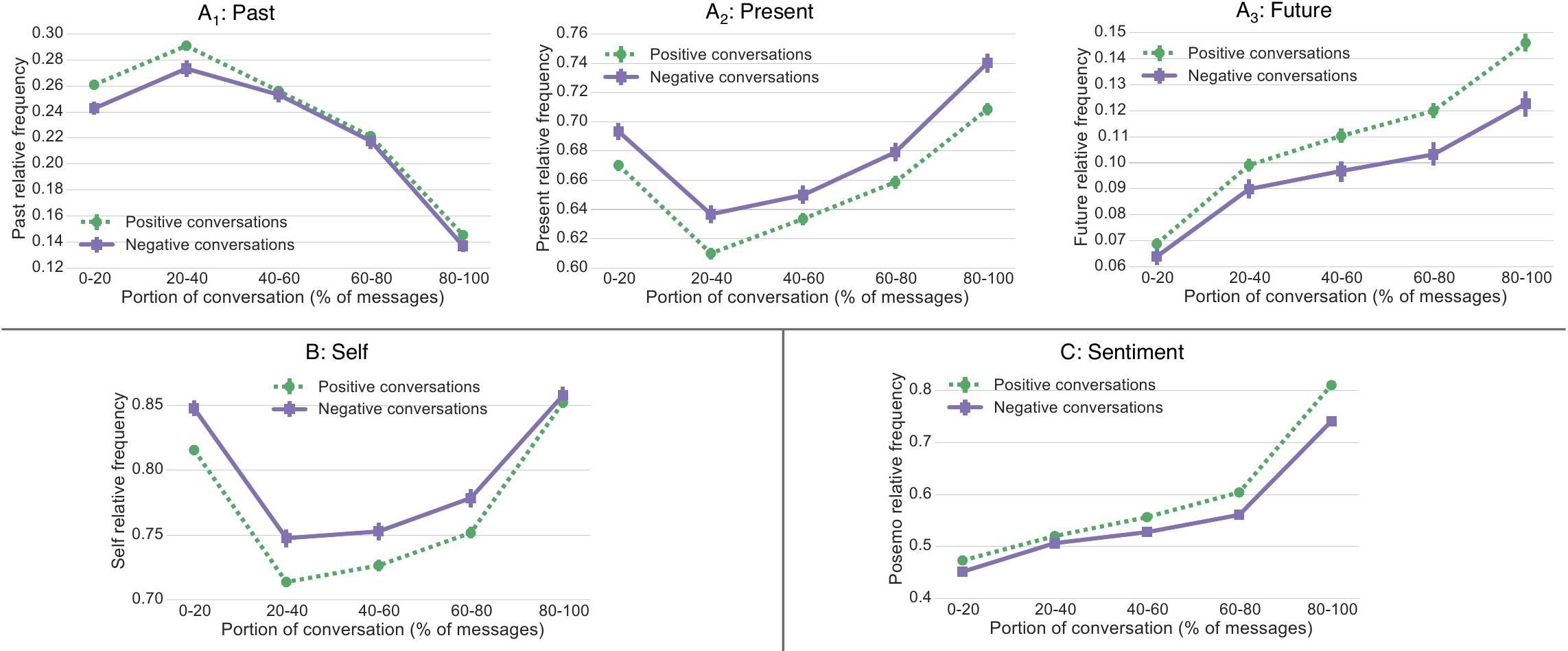}
  % \vspace{-4mm}
  \caption{A: Throughout the conversation there is a shift from talking about the past to future, where in positive conversations this shift is greater; B: Texters that talk more about others more often feel better after the conversation; C: More positive sentiment by the texter throughout the conversation is associated with successful conversations.}
  \label{fig:perspective_change_combined}
\end{figure*}

% \todo{tie it in with Beck's cognitive model~\cite{beck1967depression} and with Pyszczynski and Greenberg's self-focus model of depression~\cite{pyszczynski1987self} - see related work}
% \todo{also tie it in with pennebaker work}
Thus far, we have studied conversation dynamics and their relation to conversation success from the counselor perspective.
In this section, we show that \textit{perspective change} in the \textit{texter} over time is associated with a higher likelihood of conversation success.
Prior work has shown that day-to-day changes in writing style are associated with positive health outcomes~\cite{campbell2003secret}, and existing theories link depression to a negative view of the future~\cite{pyszczynski1987depression} and a self-focusing style~\cite{pyszczynski1987self}.
Here, we propose a novel measure to quantify three orthogonal aspects of perspective change within a single conversation: \textit{time}, \textit{self}, and \textit{sentiment}.
Further, we show that the counselor might be able to actively induce perspective change.

% In addition, we find that texters who feel better after the conversation write longer messages throughout (see Figure~\ref{fig:message_length_texter_pos_neg}).
% This suggest that conversation success could be partially ``predetermined'' by the texter. 
% \todo{Do we want to make this inference?}

\xhdr{Time} Texters start explaining their issue largely in terms of the past and present but over time talk more about the future (see Figure~\ref{fig:perspective_change_combined}A; each plot shows the relative amount of words in the LIWC past, present, and future categories~\cite{tausczik2010psychological}).
We find that texters writing more about the future are more likely to feel better after the conversation. 
This suggests that changing the perspective from issues in the past towards the future is associated with a higher likelihood of successfully working through the crisis.
% hmmm this seems like a really big stretch to me
%(consistent with the theory of Pyszczynski et al. (1987)\nocite{pyszczynski1987depression} that depression is related to a negative view of the future).

\xhdr{Self}
Another important aspect of behavior change is to what degree the texter is able to change their perspective from talking about themselves to considering others and potentially the effect of their situation on others~\cite{pyszczynski1987self,campbell2003secret}.
We measure how much the texter is focused on themselves by the relative amount of first person singular pronouns (I, me, mine) versus third person singular/plural pronouns (she, her, him / they, their), again using LIWC. %['I'], ['SheHe', 'They']
Figure~\ref{fig:perspective_change_combined}B shows that a smaller amount of self-focus is associated with more successful conversations (providing support for the self-focus model of depression~\cite{pyszczynski1987self}).
We hypothesize that the lack of difference at the end of the conversation is due to conversation norms such as thanking the counselor (``\textit{I} really appreciate it.'') even if the texter does not actually feel better.

\xhdr{Sentiment} 
Lastly, we investigate how much a change in sentiment of the texter throughout the conversation is associated with conversation success.
We measure sentiment as the relative fraction of positive words using the LIWC PosEmo and NegEmo sentiment lexicons.
The results in Figure~\ref{fig:perspective_change_combined}C show that texters always start out more negative (value below 0.5), but that the sentiment becomes more positive over time for both positive and negative conversations.
However, we find that the separation between both groups grows larger over time, which suggests that a positive perspective change throughout the conversation is related to higher likelihood of conversation success.
We find that both curves increase significantly at the very end of the conversation. 
Again, we attribute this to conversation norms such as thanking the counselor for listening even when the texter does not actually feel better.
Together with the result on talking about the future, these findings are consistent with the theory of Pyszczynski et al. (1987)\nocite{pyszczynski1987depression} that depression is related to a negative view of the future.

\xhdr{Role of a Counselor} Given that positive conversations often exhibit perspective change, a natural question is how counselors can encourage perspective change in the texter. 
We investigate this by exploring the hypothesis that the texter will tend to talk more about something (\eg, the future), if the counselor first talks about it. 
We measure this tendency using the same coordination measures as Section~\ref{subsec:coordination} except that instead of using stylistic LIWC markers (\eg, auxiliary verbs, quantifiers), we use the LIWC markers relevant to the particular aspect of perspective change (e.g., Future, HeShe, PosEmo). In all cases we find a statistically significant ($p < 0.01$; Mann-Whitney U-test) increase in the likelihood of the texter using a LIWC marker if the counselor used it in the previous message (\texttildelow4-5\% change).
%: $C^{\text{Future}}(\text{texter, counselor}) = 0.0539$, $C^{\text{HeShe}}(\text{texter, counselor}) = 0.0388$, $C^{\text{PosEmo}}(\text{texter, counselor}) = 0.0389$. 
% While one might expect this result due to conversation dynamics~\cite{Danescu-Niculescu-Mizil:thesis2012}, this does suggest a link between perspective change and how the counselor conducts the conversation. %These findings are consistent with results in psychology \cite{}. 
This link between perspective change and how the counselor conducts the conversation suggests that the counselor might be able to actively induce measurable perspective change in the texter.

%% figures
% \begin{figure}[t]
%   \centering
%   \includegraphics[width=1.0\linewidth]{message_length_texter_pos_neg.pdf}
%   \caption{Texters that feel better after the conversation write longer messages throughout.}
%   \label{fig:message_length_texter_pos_neg}
% \end{figure}

%\begin{figure*}[t]
%  \centering
%  \includegraphics[width=.32\linewidth]{past_texter_pos_neg.pdf}
%  \includegraphics[width=.32\linewidth]{present_texter_pos_neg.pdf}
%  \includegraphics[width=.32\linewidth]{future_texter_pos_neg.pdf}
%  \caption{Throughout the conversation there is a shift from past to present and future.
%           In particular, talking more about the future is associated with successful conversations.
%           \todo{this could be compressed into one figure. or just use future vs rest.}
%  }
%  \label{fig:time_texter_pos_neg}
%\end{figure*}
%
%\begin{figure}[t]
%  \centering
%  \includegraphics[width=1.0\linewidth]{self_vs_other_texter_pos_neg.pdf}
%  \caption{Texter that talk more about others more often feel better after the conversation.}
%  \label{fig:self_vs_other_texter_pos_neg}
%\end{figure}
%
%\begin{figure}[t]
%  \centering
%  \includegraphics[width=1.0\linewidth]{positive_texter_pos_neg.pdf}
%  \caption{More positive sentiment by the texter throughout the conversation is associated with successful conversation.}
%  \label{fig:positive_texter_pos_neg}
%\end{figure}

\section{Predicting Counseling Success}
\label{sec:prediction}
% !TEX root = paper-counseling.tex

In this section, we combine our quantitative insights into a prediction task. 
We show that the linguistic aspects of crisis counseling explored in previous sections have predictive power at the level of individual conversations by evaluating their effectiveness as features in classifying the outcome of conversations. % as positive or negative. 
Specifically, we create a balanced dataset of positive and negative conversations more than 30 messages long 
% mention min length=30?
and train a logistic regression model to predict the outcome given the first $x$\% of messages in the conversation. %Because our focus is in on finding effective counseling strategies, we focus on features from the counselor messages for this task. 
There are 3619 such negative conversations and 
and we randomly subsample the larger set of positive conversations.
% we chose the same number of positive conversations at random to add to the dataset.
We train the model with batch gradient descent and use L1 regularization when n-gram features are present and L2 regularization otherwise. 
We evaluate our model with 10-fold cross-validation and compare models using the area under the ROC curve (AUC). 

%Unsurprisingly, the model's accuracy increases rapidly with $x$, but we also find the model does quite well after only being exposed to the first 10\% of a conversation (see Figure ?). We attribute the significant increase in performance for $x = 100$ to strong linguistic cues that appear as a conversation wraps up (e.g., ``I'm glad this conversation helped you feel better''). 
%% Could perhaps mention the interesting observation that some users are polite and sound positive even in negative conversations here
%To avoid this issue while still performing our evaluation on as much data as possible, our detailed evaluation of specific features is performed with $x = 80$. 

%In our feature analysis, we explore the three aspects of perspective change discussed in section ?. In particular, we add use two sentiment features: the relative fraction of positive words using the LIWC PosEmo and NegEmo sentiment lexicons and the scores produced by the VADER sentiment analysis tool \cite{}, three time features indicating the relative frequencies of LIWC past, future, and present words, and one self feature measuring the relative amount of first person singular pronouns versus third person singular and plural pronoun, also using LIWC.

\xhdr{Features}
We include three aspects of counselor messages discussed in Section~\ref{sec:ambiguity}: hedges, check questions, and the similarity between the counselor's message and previous texter message. 
%For the first two, we find occurrences of check questions and hedges with regular expressions and use the frequency of occurrence per message as features. For the third, we use the average Jaccard similarity between the counselor's message and the previous texter message. 
We add a measure of how much progress the counselor has made (Section~\ref{sec:conversation_progress}) by computing the Viterbi path of stages for the conversation (only for the first $x$\%) with the HMM conversation model and then adding the duration of each stage (in \#messages) as a feature. 
Additionally, we add average message length and average sentiment per message using VADER sentiment~\cite{hutto2014vader}. 
Further, we add temporal dynamics to the model by adding feature conjunctions with the stages HMM model. After running the stages model over the $x$\% of the conversation available to the classifier, we add each feature's average value over each stage as additional features. 
% For example the feature {\sc message\_length\_2} would represent the average message length during stage 2.
Lastly, we explore the benefits of adding surface-level text features to the model by adding unigram and bigram features. 
Because the focus of this work is on counseling strategies, we primarily experiment with models using only features from counselor messages. 
For completeness, we also report results for a model including texter features. % (see last row of Table~\ref{tab:prediction}).
%we also report the score when texter bigrams are included for completeness.

 %The common approach of using n-gram counts for this adds thousands of sparse features to the model, complicating the feature analysis. Instead, we add a single feature representing the difference in log likelihoods assigned to the conversation by the KenLM language model model \cite{heafield2011kenlm} trained on positive conversations and a language model trained on negative ones. 
% Is it worth going into the details in having to do cross validation here?

 \begin{figure}[t]
  \centering
  \includegraphics[width=1.0\linewidth]{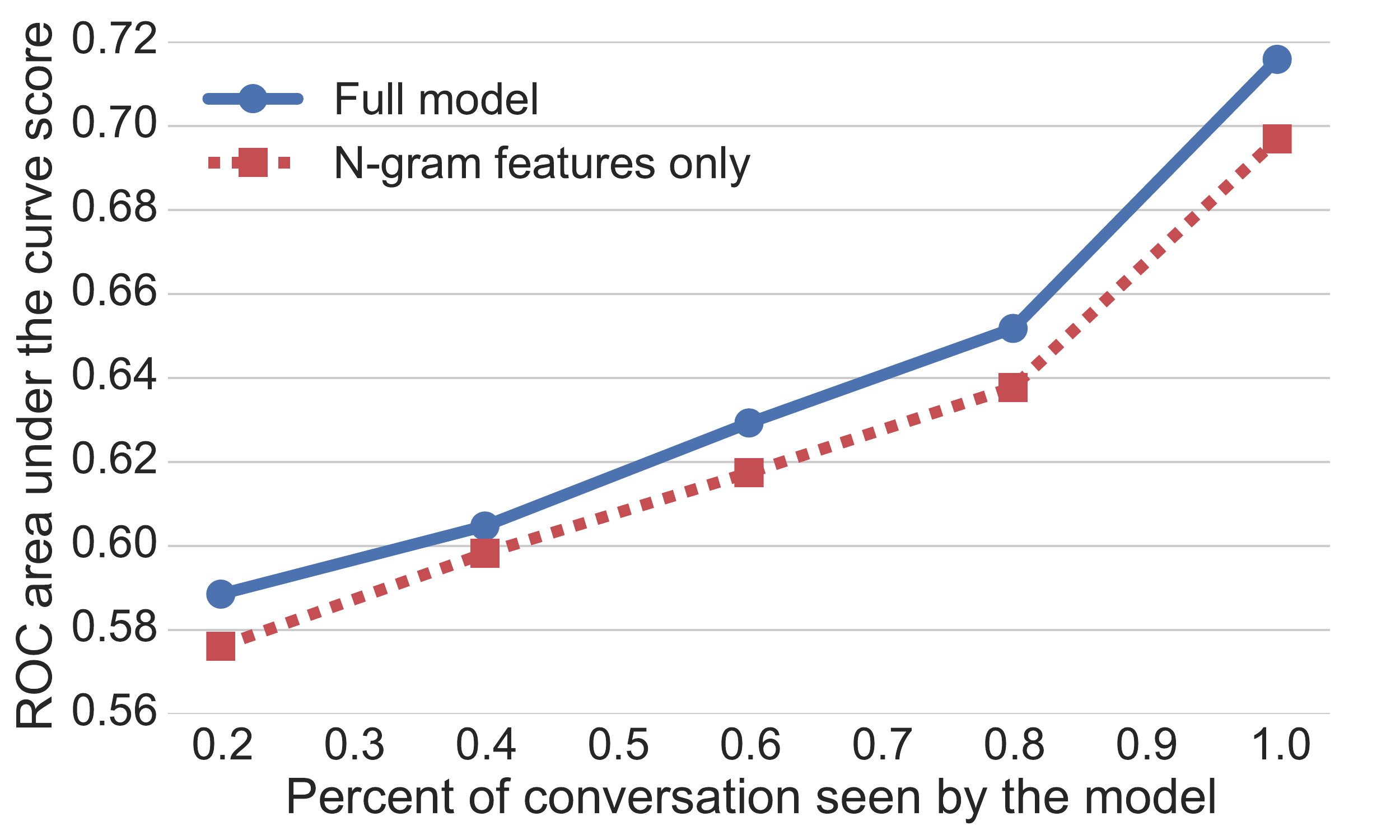}
   \vspace{-5mm}
  \caption{Prediction accuracies vs. percent of the conversation seen by the model (without texter features).}
  \label{fig:prediction_accuracies}
\end{figure}

\xhdr{Prediction Results}
The model's accuracy increases with $x$, and we show that the model is able to distinguish positive and negative conversations after only seeing the first 20\% of the conversation (see Figure~\ref{fig:prediction_accuracies}). 
We attribute the significant increase in performance for $x = 100$ (Accuracy=0.687, AUC=0.716) to strong linguistic cues that appear as a conversation wraps up (e.g., ``I'm glad you feel better.''). 
%% Could perhaps mention the interesting observation that some users are polite and sound positive even in negative conversations here
To avoid this issue, our detailed feature analysis is performed at $x = 80$. 
%while still performing our evaluation on as much data as possible

% ROC seems more reliable of a metric, maybe should put it first?
% include deltas? might not be enough space to include both deltas and accuracy
% maybe reorder: first stages then text and swtitch to n-gram features
\begin{table}[t]
\centering
\small % \footnotesize
\resizebox{1.0\columnwidth}{!}{%
\begin{tabular}{ll}
  %\hline
  Features &  ROC AUC \\ \hline
  Counselor unigrams only & 0.630 \\
  Counselor unigrams and bigrams only & 0.638 \\
  None   &  0.5     \\    
  + hedges &  0.514 (+0.014) \\ 
  + check questions & 0.546 (+0.032)  \\ 
  + similarity to last message & 0.553 (+0.007)  \\ 
  + duration of each stage & 0.561 (+0.008)  \\  
  + sentiment & 0.590 (+0.029)  \\ 
  + message length & 0.596 (+0.006) \\ 
  + stages feature conjunction & 0.606 (+0.010) \\ 
  + counselor unigrams and bigrams & \textbf{0.652 (+0.046)}  \\
  + texter unigrams and bigrams & \textbf{0.708 (+0.056)}  \\ %\hline
 \end{tabular}
 }
 % \vspace{-3mm}
 \caption{
 % Conversation outcome prediction performance of nested models.
 Performance of nested models predicting conversation outcome given the first $80$\% of the conversation. % using ROC AUC measure. 
 In bold: full models with only counselor features and with additional texter features. 
 % \jure{Why not add Bag of words baseline here? Does it perform too well? Too badly? }
 }
 \label{tab:prediction}
 \end{table}

% \pagebreak[4]
\xhdr{Feature Analysis}
The model performance as features are incrementally added to the model is shown in Table~\ref{tab:prediction}. 
All features improve model accuracy significantly ($p < 0.001$; paired bootstrap resampling test). 
Adding n-gram features produces the largest boost in AUC and significantly improves over a model just using n-gram features (0.638 vs. 0.652 AUC).
Note that most features in the full model are based on word frequency counts that can be derived from n-grams which explains why a simple n-gram model already performs quite well. 
% which is unsurprising because most of the other features are based on word frequency counts that can be derived from n-grams. 
However, our model performs well with only a small set of linguistic features, demonstrating they provide a substantial amount of the predictive power.
%Although adding n-gram features produces the largest boost in AUC, the model already performs well with the small set of linguistic feature demonstrating they provide much of the predictive power. 
The effectiveness of these features shows that, in addition to exhibiting group-level differences reported earlier in this paper, they provide useful signal for predicting the outcome of individual conversations.

\section{Conclusion \& Future Work}
\label{sec:conclusion}
% !TEX root = paper-counseling.tex

Knowledge about how to conduct a successful counseling conversation has been limited by the fact that studies have remained largely qualitative and small-scale.
In this work, we presented a large-scale quantitative study on the discourse of counseling conversations.
We developed a set of novel computational discourse analysis methods suited for large-scale datasets and used them to discover actionable conversation strategies that are associated with better conversation outcomes.
We hope that this work will inspire future generations of tools available to people in crisis as well as their counselors.
% For example, the insights gained through this work could be helpful from improving counselor training and making them more effective at supporting people in crisis to real-time counseling quality monitoring and automated answer suggestion support tools.
For example, our insights could help improve counselor training 
and give rise to real-time counseling quality monitoring and answer suggestion support tools.
% provide a basis for 

% \begin{itemize}
%   \item matching conversation trajectory, e.g. through sentence embedding models specifically trained in this context?
%   \item automated question answering or answer suggestion, e.g. through machine translation approach
%   \item Abstract vs concrete (have tried using~\cite{brysbaert2014concreteness} but found it to be very correlated with length)
%   \item Formal vs informal language
%   \item probably not(Sharing a secret (has been shown helpful in Pennebakers work), hard to classify automatically)
%   \item ...
% \end{itemize}

% Acknowledgments:
\xhdr{Acknowledgements}
We thank Bob Filbin for facilitating the research, 
Cristian Danescu-Niculescu-Mizil for many helpful discussions, 
and Dan Jurafsky, Chris Manning, Justin Cheng, Peter Clark, David Hallac, Caroline Suen, Yilun Wang and the anonymous reviewers for their valuable feedback on the manuscript. 
%
% This research has been supported in part by the Mobilize Center (NIH U54 EB020405).
% \todo{Grants from Jure}
This research has been supported in part by NSF
CNS-1010921,              % NSF with Madhav (Sept 2015)
IIS-1149837,                % NSF CAREER (Dec 2015)
NIH BD2K,     % Mobilize
ARO MURI,                 % 
DARPA XDATA,              % 
DARPA SIMPLEX,            % 
Stanford Data Science Initiative,
Boeing,                   % Oct 2016
Lightspeed,          % Jan 2017
SAP,                      % Jan 2017
and Volkswagen.           % Jan 2017

% \vspace{-\baselineskip}

\hyphenation{LePendu}
\balance
\bibliographystyle{acl2012}
\bibliography{refs}

% \pagebreak
% \appendix
% \section{Appendix}
% \label{sec:appendix}
% \input{110appendix}

\end{document}